\documentclass[nohyperref]{article}

\usepackage{microtype}
\usepackage{graphicx}
\usepackage{subfigure}
\usepackage{booktabs} % for professional tables

\usepackage{hyperref}

\usepackage[accepted]{icml2022}

\usepackage{amsmath}
\usepackage{amssymb}
\usepackage{mathtools}
\usepackage{amsthm}

\usepackage[capitalize,noabbrev]{cleveref}

\definecolor{mine}{RGB}{205, 232, 248}%

\theoremstyle{plain}
\newtheorem{theorem}{Theorem}[section]
\newtheorem{proposition}[theorem]{Proposition}

\newtheorem{corollary}[theorem]{Corollary}
\theoremstyle{definition}

\theoremstyle{remark}

\usepackage[textsize=tiny]{todonotes}
\usepackage{wrapfig}
\usepackage{enumitem}
\usepackage{enumerate}
\usepackage{multicol}
\usepackage{multirow}
\usepackage{algorithm}
\usepackage{algorithmic}
\usepackage{amsmath, amsthm, amssymb, amsfonts}
\usepackage{dsfont}
\usepackage{lipsum}
\usepackage{nicefrac}
\usepackage{bm}
\usepackage{float}
\icmltitlerunning{Discriminator-Weighted Behavioral Cloning}

\begin{document}

\twocolumn[
\icmltitle{Discriminator-Weighted Offline Imitation Learning \\ from Suboptimal Demonstrations}
% \icmltitle{Cooperation-Oriented Offline Imitation Learning \\  with Supplementary Unlabeled Experience}



\icmlsetsymbol{equal}{*}

\begin{icmlauthorlist}
\icmlauthor{Haoran Xu}{jd}
\icmlauthor{Xianyuan Zhan}{tsinghua}
\icmlauthor{Honglei Yin}{jd}
\icmlauthor{Huiling Qin}{jd}
\end{icmlauthorlist}

\icmlaffiliation{jd}{JD Technology, Beijing, China}
\icmlaffiliation{tsinghua}{Institute for AI Industry Research (AIR), Tsinghua University, Beijing, China}

\icmlcorrespondingauthor{Haoran Xu}{ryanxhr@gmail.com}
\icmlcorrespondingauthor{Xianyuan Zhan}{zhanxianyuan@air.tsinghua.edu.cn}

\icmlkeywords{Imitation Learning, Offline Imitation Learning, Offline Reinforcement Learning}

\vskip 0.3in
]



\printAffiliationsAndNotice{}  % leave blank if no need to mention equal contribution

\begin{abstract}
We study the problem of offline Imitation Learning (IL) where an agent aims to learn an optimal expert behavior policy without additional online environment interactions. Instead, the agent is provided with a supplementary offline dataset from suboptimal behaviors. 
Prior works that address this problem either require that expert data occupies the majority proportion of the offline dataset, or need to learn a reward function and perform offline reinforcement learning (RL) afterwards. In this paper, we aim to address the problem without additional steps of reward learning and offline RL training for the case when demonstrations contain a large proportion of suboptimal data. 
Built upon behavioral cloning (BC), we introduce an additional discriminator to distinguish expert and non-expert data. We propose a cooperation framework to boost the learning of both tasks, Based on this framework, we design a new IL algorithm, where the outputs of discriminator serve as the weights of the BC loss. Experimental results show that our proposed algorithm achieves higher returns and faster training speed compared to baseline algorithms. Code is available at \url{https://github.com/ryanxhr/DWBC}.
\end{abstract}

\section{Introduction}
The recent success of reinforcement learning (RL) in many domains showcases the great potential of applying this family of learning methods to real-world applications. 
A key prerequisite for RL is to design a reward function that specifies what kind of agent behavior is preferred. However, in many real-world applications, designing a reward function is prohibitively difficult~\citep{ng1999policy,rlblogpost}.
By contrast, imitation learning (IL) provides a much easier way to leverage the reward function implicitly from the collected demonstrations and has achieved great success in many sequential decision making problems~\citep{pomerleau1989alvinn,ng2000algorithms,ho2016generative}.

However, popular IL methods such as behavioral cloning (BC)~\citep{pomerleau1989alvinn} and generative adversarial imitation learning (GAIL)~\citep{ho2016generative} assume the expert demonstration is optimal. 
Unfortunately, it is often difficult to obtain sufficient optimal demonstrations for many real-world tasks, because human experts often make mistakes due to various reasons, such as the difficulty of the task, partial observability of the environment, or the presence of distraction. 
Given such noisy expert demonstrations, which contain records of both optimal and non-optimal behaviors, BC and GAIL all fail to imitate the optimal policy~\citep{wu2019imitation,ma2020adversarial}.
Current methods that deal with suboptimal demonstrations either require additional labels, which can be done explicitly by annotating each demonstration with confidence scores by human experts~\citep{wu2019imitation}, or implicitly by ranking noisy demonstrations according to their relative performance through interacting with the environment~\citep{brown2019extrapolating,brown2020better,zhang2021confidence}.
However, human annotation and environment interaction are laborious and expensive in real-world settings, such as in medicine, healthcare, and industrial processes.

In this work, we investigate a pure offline learning setting where the agent has access to neither the expert nor the environment for additional information. The agent, instead, has only access to a small pre-collected dataset sampled from the expert and a large batch offline dataset sampled from one or multiple behavior policies that could be highly sub-optimal. 
This strictly offline imitation learning problem arises in many real-world problems, where environment interactions and expert annotations are costly. 
Prior works that address the problem are based on variants of BC or inverse RL. \citet{sasaki2021behavioral} reuses another policy learned by BC as the weight of original BC objective. However, this requires that expert data occupy the majority proportion of the offline dataset, otherwise the policy will be misguided to imitate the suboptimal data. \citet{zolna2020offline} first learns a reward function that prioritizes expert data over others and then performs offline RL based on this reward function. This algorithm is extremely expensive to run, requiring solving offline RL in an inner loop, which itself is a challenging problem and prone to training instability~\citep{kumar2019stabilizing} and hyperparameter sensitivity~\citep{wu2019behavior}.

In this paper, we propose an offline imitation learning algorithm to learn from demonstrations that (perhaps) contain a large proportion of suboptimal data without additional steps of reward learning and offline RL training. 
Built upon the task of BC, we introduce an additional task to learn a discriminator to distinguish expert and non-expert data. 
We propose a cooperation framework to learn the policy and discriminator cooperatively and boost the performance of both tasks. 
Based on this framework, we adopt a worst-case error minimization strategy to the policy such that the discriminator can be more robustly learned.
This results in a new offline policy learning objective, and surprisingly, we find its equivalence to a generalized BC objective, where the outputs of the discriminator serve as the weights of the BC loss function. 
We thus term our resulting algorithm Discriminator-Weighted Behavioral Cloning (DWBC). 
Experimental results show that DWBC achieves higher returns and faster training speed compared to baseline algorithms under different scenarios.

To summarize, the contributions of this paper are as follows.
\begin{itemize}[leftmargin=*,nosep]
\item We propose a cooperation framework to learn the policy and discriminator cooperatively and boost the performance of both tasks (Section~\ref{sec:cooperative});
\item Based on the proposed framework, we design an effective and light-weighted offline IL algorithm with a worst-case error minimization strategy (Section~\ref{sec:dwbc});
\item We present promising comparison results with comprehensive analysis for our algorithm, which surpasses the state-of-the-art methods (Section~\ref{sec:comp_eval});
\item As a by-product, we show that the discriminator in our algorithm can be used to perform offline policy selection, which is of independent interest (Section~\ref{sec:ops}).
\end{itemize}

\section{Preliminary}
\subsection{Problem Setting}
We consider the standard fully observed Markov Decision Process (MDP) setting~\citep{sutton1998introduction}, $\mathcal{M}=\left\{\mathcal{S}, \mathcal{A}, P, r, \gamma, d_{0}\right\}$, where $\mathcal{S}$ is the state space, $\mathcal{A}$ is the action space, $P: \mathcal{S} \times \mathcal{A} \rightarrow \Delta(\mathcal{S})$ is the MDP's transition probability, $r$ is the reward function, $\gamma \in [0,1)$ is the discount factor for future reward and $d_{0}$ is the initial distribution. A policy $\pi: \mathcal{S} \rightarrow \Delta(\mathcal{A})$ maps from state to distribution over actions. We denote $d^{\pi} \in \Delta(\mathcal{S} \times \mathcal{A})$ as the discounted state-action distribution of $\pi$ under transition kernel $P$, that is, $d^{\pi}=(1-\gamma) \sum_{t=0}^{\infty} \gamma^{t} d_{t}^{\pi}$, where $d_{t}^{\pi} \in \Delta(\mathcal{S} \times \mathcal{A})$ is the distribution of $\left(s^{(t)}, a^{(t)}\right)$ under $\pi$ at step $t$. 
% Given a reward function $f: \mathcal{S} \times \mathcal{A} \mapsto[0,1], V_{P, f}^{\pi}$ denotes the expected cumulative cost of $\pi$ under the transition kernel $P$ and cost function $f$. 
Following the standard IL setting, the ground truth reward function $r$ is unknown. Instead, we have the demonstrations collected by the expert denoted as $\pi_{e}: \mathcal{S} \rightarrow \Delta(\mathcal{A})$ (potentially stochastic and not necessarily optimal). Concretely, we have an expert dataset in the form of i.i.d tuples $\mathcal{D}_{e}=\left\{s_{i}, a_{i}, s_{i}^{\prime}\right\}_{i=1}^{n_{e}}$ where $(s, a)$ is sampled from distribution $d^{\pi_{e}}$ and $s^{\prime}$ is sampled from $P(s, a)$.

% Let $\mathcal{D}_e$ denote the expert dataset, $\mathcal{D}_n$ denote the non-expert dataset, $\mathcal{D}_b=\mathcal{D}_e \cup \mathcal{D}_n$ denote the batch dataset. We assume that the expert dataset may contain non-expert data and the non-expert dataset may contain expert data.
In our problem setting, we also have an offline static dataset consisting of i.i.d tuples $\mathcal{D}_{o}=\left\{s_{i}, a_{i}, s_{i}^{\prime}\right\}_{i=1}^{n_{o}} \ \text{s.t.} \ (s, a) \sim \rho(s, a), s^{\prime} \sim P(s, a)$, where $\rho \in \Delta(\mathcal{S} \times \mathcal{A})$ is an offline state-action distribution resulting from some other behavior policies. Note that these behavior policies could be much worse than the expert $\pi_{e}$.
Our goal is to only leverage the offline batch data $\mathcal{D}_b=\mathcal{D}_e \cup \mathcal{D}_o$ to learn an optimal policy $\pi$ with regard to optimizing the ground truth reward $r$, without any interaction with the environment or the expert.
% More specifically, our goal is to utilize the offline static data $\mathcal{D}_{o}$ to combat covariate shift and learn a policy that can significantly outperform traditional offline IL methods such as Behavior cloning (BC), without any interaction with the real world or expert.

\subsection{A Generalized Behavioral Cloning Objective}
\label{sec_general_bc}
In order to discard low-quality demonstrations and only clone the best behavior available, we consider a generalized behavioral cloning objective to imitate demonstrations unequally, that is,
\begin{equation}
\label{eq_general_bc}
\min_{\pi} \underset{(s, a) \sim \mathcal{D}_{b}}{\mathbb{E}} \left[-\log \pi(a|s) \cdot f(s,a) \right],
\end{equation}
where $f: \mathcal{S} \times \mathcal{A} \rightarrow[0,1]$ denotes an arbitrary weight function. 
Existing offline IL methods can simply be recovered by picking one of the valid weight configurations:
\begin{itemize}[leftmargin=*,nosep]
\item If $f(s, a)=1$ for $\forall(s, a) \in \mathcal{S} \times \mathcal{A}$, the objective~(\ref{eq_general_bc}) corresponds to the vanilla BC objective. 
\item If $f(s, a)=\pi^{\prime}(a|s)$, where $\pi^{\prime}$ is an old policy which was previously optimized with $\mathcal{D}_{b}$, the objective~(\ref{eq_general_bc}) corresponds to the objective of Behavioral Cloning from Noisy Demonstrations~\citep{sasaki2021behavioral}.
Since $\sum_{a} \pi^{\prime}(a|s)=1$ for $\forall s \in \mathcal{S}$ is satisfied, $\pi^{\prime}(a|s)$ can be interpreted as the weights for weighted action sampling.
\item If $f(s, a)=\mathds{1} \left[ A^{\pi}(s, a) \right]$, where $\mathds{1}$ is the indicator function which creates a boolean mask that eliminates samples which are thought to be worse than the current policy, the objective~(\ref{eq_general_bc}) corresponds to the objective of Offline Reinforced Imitation Learning~\citep{zolna2020offline}.
\end{itemize}

The objective~(\ref{eq_general_bc}) can also be deemed as the objective of Soft Q Imitation Learning~\citep{reddy2019sqil} with $f(s,a)=1$ for $(s,a) \in \mathcal{D}_e$ and $f(s,a)=0$ for $(s,a) \in \mathcal{D}_o$ in online IL literature; or the objective of off-policy actor-critic (Off-PAC) algorithm~\citep{degris2012off} with $f(s,a)=Q^{\pi}(s,a) \cdot \pi(a|s)/\pi_{b}(a|s)$ in online RL literature.

\begin{figure*}[t]
\centering
\includegraphics[width=1.8\columnwidth]{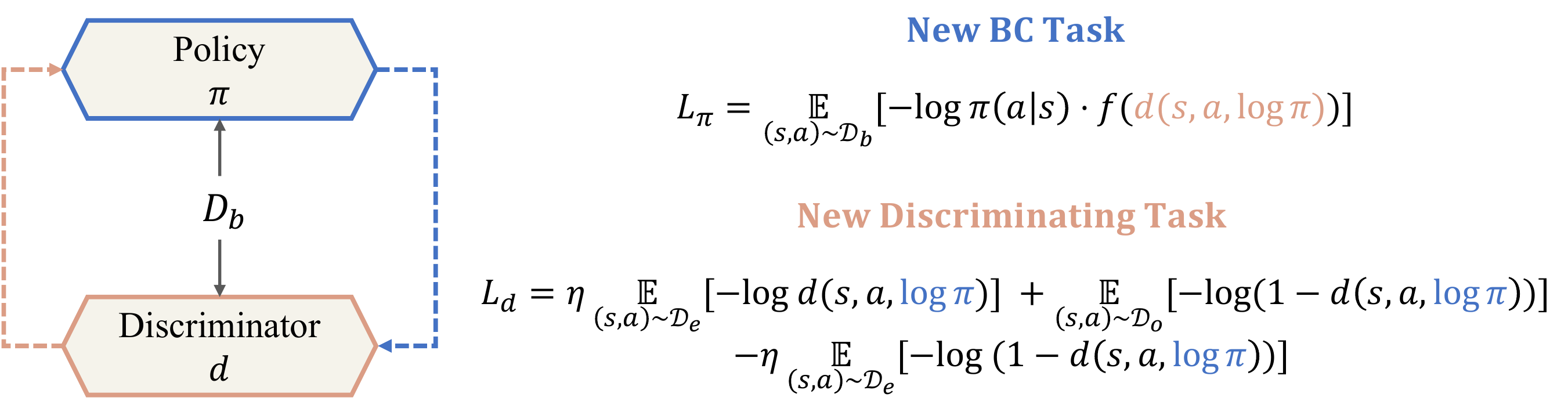} 
\caption{Illustration of our proposed cooperation framework to alternately learn $\pi$ and $d$. In this framework, $\pi$ uses the outputs of $d$ as the weights to perform a new BC Task; $d$ includes $\pi$ as additional input to form a new Discriminating Task. This framework is different from GAN-style frameworks in that: 1) $\pi$ and $d$ are learned cooperatively rather than adversarially; 2) the training of $\pi$ and $d$ are decoupled into individual objectives rather than sharing one coupled objective.}
\label{fig_illustration}
\end{figure*}

\section{Methodology}
We now continue to describe our approach for offline imitation learning from demonstrations that (perhaps) contain large-proportional suboptimal data, without additional steps of reward learning and offline RL training. 
Built upon the task of BC, we introduce an additional task to learn a discriminator to distinguish expert and non-expert data. 
We propose a cooperation framework to boost the performance of both tasks. 
Based on this framework, we adopt a worst-case error minimization strategy to the policy such that the discriminator can be more robustly learned.
This results in a new generalized BC objective, we then provide the interpretation of weights in our generalized BC objective, this gives the intuition about why our method can work.

\subsection{Learn the Policy and Discriminator Separately}
It is obvious that we can avoid the negative impact of suboptimal demonstrations presented in $\mathcal{D}_o$ by only imitating $\mathcal{D}_e$, which can be written as 
\begin{equation}
\label{eq_task_bc}
\min_{\pi} \underset{(s, a) \sim \mathcal{D}_{e}}{\mathbb{E}} \left[-\log \pi(a|s) \right].
\end{equation}

We call the task of learning a policy using objective~(\ref{eq_task_bc}) as \textbf{BC Task}. The drawback of BC task is that it does not fully utilize the information from $\mathcal{D}_o$, the resulting policy may not be able to generalize and will suffer from compounding errors due to the potential limited size and state coverage of $\mathcal{D}_e$~\citep{ross2011reduction}. If we can select those high-reward transitions from $\mathcal{D}_o$ and combine them with $\mathcal{D}_e$, we are expected to get a better policy.

Now let's consider another different task, which aims to learn a discriminator by contrasting expert and non-expert transitions, given by
\begin{equation}
\label{eq_task_d}
\min_{d} \underset{(s, a) \sim \mathcal{D}_{e}}{\mathbb{E}} \left[-\log d(s,a) \right] + \underset{(s, a) \sim \mathcal{D}_{o}}{\mathbb{E}} \left[-\log(1-d(s,a)) \right].
\end{equation}

Objective (\ref{eq_task_d}) is similar to how the discriminator is trained in GAIL~\citep{ho2016generative} and GAN~\citep{goodfellow2014generative}, except that the second term is sampled from a fixed dataset instead of new samples drawn from the learned policy by interacting with the environment.

However, optimizing objective~(\ref{eq_task_d}) will make the learned discriminator assign 1 to all transitions from $\mathcal{D}_e$ and 0 to all transitions from $\mathcal{D}_o$. This limiting behavior is unsatisfactory because $\mathcal{D}_o$ can contain some successful (high-reward) transitions. This bears similarity to the positive-unlabeled classification problem~\citep{elkan2008learning}, where both positive and negative samples exist in the unlabeled data. 

To solve this problem, previous works adopt the approach from positive-unlabeled (PU) learning~\citep{du2015convex,xu2019positive,zolna2020combating}. The main idea is to re-weight different losses for positive and unlabeled data, in order to obtain an estimate of model loss on negative samples that is not directly available. Applying PU learning to objective~(\ref{eq_task_d}) yields the following objective:
\begin{align}
\min_{d}\; &\eta \underset{(s, a) \sim \mathcal{D}_{e}}{\mathbb{E}} \left[-\log d(s,a) \right] \notag\\
&+ \underset{(s, a) \sim \mathcal{D}_{o}}{\mathbb{E}} \left[-\log(1-d(s,a)) \right] \notag\\
&- \eta \underset{(s, a) \sim \mathcal{D}_{e}}{\mathbb{E}} \left[-\log(1-d(s,a)) \right], \label{eq_task_d_pu}
\end{align}
where $\eta$ is a hyperparameter, corresponds to the proportion of positive samples to unlabeled samples.
We call the task of learning a discriminator using objective~(\ref{eq_task_d_pu}) as \textbf{Discriminating Task}.
Intuitively, the second term in~(\ref{eq_task_d_pu}) could make $d(s,a)$ of state-action pairs from $\mathcal{D}_e$ become 0 if similar state-action pairs are included in $\mathcal{D}_o$, and the third term in~(\ref{eq_task_d_pu}) balances the impact of the second term, i.e., avoids $d(s,a)$ of state-action pairs from $\mathcal{D}_e$ becoming 0.

However, using information from state and action may be insufficient for the learning of Discriminating task.
For example, suppose $\mathcal{D}_{e}$ comprises near-end transitions of expert trajectories, whereas $\mathcal{D}_{o}$ comprises near-front transitions of expert trajectories and transitions from non-expert trajectories. 
% For example, suppose $\mathcal{D}_{e}$ comprises less successful transitions of expert trajectories, whereas $\mathcal{D}_{o}$ comprises some successful transitions of expert trajectories and transitions from non-expert trajectories. 
In this case, it is hard for the discriminator to distinguish between expert transitions and non-expert transitions in $\mathcal{D}_{o}$, as states of transitions in $\mathcal{D}_{o}$ bear a large similarity (these states are near the initial state), but have a large difference from states in $\mathcal{D}_{e}$. 

To summarize, BC Task aims to imitate the expert behavior from $\mathcal{D}_e$, but ignores the valuable information from $\mathcal{D}_o$; Discriminating Task aims to contrast expert and non-expert transitions from $\mathcal{D}_e$ and $\mathcal{D}_o$, but only uses state-action information as input. 
Both tasks lack enough information to improve their own performance, which however, can be obtained from the other task, as we will elaborate next.
% It then seems natural to find a scheme to incorporate the policy into the training of the discriminator and effectively use the discriminator to help the training of the policy.

\subsection{Learn the Policy and Discriminator Cooperatively}\label{sec:cooperative}
We propose a cooperation framework to learn the policy and discriminator cooperatively. 
In this framework, we aim to boost the performance of BC Task and Discriminating Task by incorporating the policy into the training of the discriminator and effectively using the discriminator to help the training of the policy.
As illustrated in Figure \ref{fig_illustration}, the policy $\pi$ uses the discriminator $d$ to perform a new BC Task (i.e., generalized behavioral cloning as introduced in Section 2.2), where the weight is a function of $d$. The discriminator $d$ also gets information from the policy $\pi$ as additional input, yielding a new Discriminating Task.

Suppose that $d$ is well-learned to be able to contrast expert and non-expert transitions in $\mathcal{D}_b=\mathcal{D}_e \cup \mathcal{D}_o$, the policy will become better if one can choose an appropriate weight function $f$ to make $\pi$ only imitate the expert data in $\mathcal{D}_b$. By this way, we are able to use the entire dataset $\mathcal{D}_b$ but get rid of the negative impact of those low-quality data.

Supposed $\pi$ is learned to be optimal, i.e., assigns large probabilities to expert actions in expert states, the discriminator will receive additional learning signal. It will be easier for the discriminator to contrast expert and non-expert transitions in $\mathcal{D}_o$, as $\pi(a|s)$ will be large if $(s,a)$ are from expert behaviors and small if $(s,a)$ are from non-expert behaviors. Without this information from $\pi$, the discriminator is much harder to learn by only using information from $(s,a)$.

A keen reader may find the similarity of our propased framework and GAN-style frameworks~\citep{goodfellow2014generative,ho2016generative}, where the policy and the discriminator are also jointly learned. 
However, the learning strategy of our framework has several differences compared with GAN-style frameworks. 
In GAN, the policy aims to generate expert data and the discriminator aims to distinguish between expert data and policy generated data. If the policy perfectly matches the expert, the discriminator will be unable to distinguish well, and vice versa. This means that GAN adopts an adversarial framework, where task A and task B are contradictory to each other, an improved performance of one task will lead to a deteriorated performance of another task.
In contrast to adversarial, our framework is cooperative, task A and task B cooperate with each other to help both tasks, an improved performance of one task will also lead to an improved performance of another task.

Moreover, GAN-style frameworks need to solve a min-max optimization problem (i.e., $\min_{d}\max_{\pi}\mathcal{L}(d, \pi)$) and is known to suffer from issues such as traning instability and mode collapse~\citep{arjovsky2017wasserstein}.
Whereas our framewwork allows the decoupled training of $\pi$ and $d$. They can both learn with their own objectives in a fully supervised manner (see Figure \ref{fig_illustration}), which is very easy to train and computationally cheap.

\subsection{Discriminator-Weighted Behavioral Cloning}\label{sec:dwbc}
% \subsection{Cooperation-Oriented Offline Imitation Learning}\label{sec:dwbc}
It is obvious that, in our proposed framework, there exists multiple valid choices of weight function $f$ that can make the policy imitate those high-reward transitions in $\mathcal{D}_{b}$. For example, $f$ could be $\mathds{1}[d>0.5]$ or $\exp(d/\beta)$, where $\beta > 0$ is a hyperparameter and $\mathds{1}$ is the indicator function. 
However, does there exsit one principled solution of $f$?

Notice that now $\pi$ appears in the input of $d$, this means that imitation information from $\log\pi$ will affect $\mathcal{L}_d$, and further impact the learning of $d$. Hence both $d$ and $\mathcal{L}_d$ become functionals of $\pi$ (function of a function), i.e., $d(s,a,\log\pi(a|s))$ and $\mathcal{L}_d(d,\log\pi)$.
Inspired by the idea of adversarial training, we make the policy $\pi$ challenge the discriminator $d$ by doing the opposite to minimizing $\mathcal{L}_d$, in other words, we let $\pi$ maximize $\mathcal{L}_d$ under current $d$.
This can be seen as minimizing the worst-case error~\citep{carlini2019evaluating,fawzi2016robustness,goodfellow2014explaining}, which makes the robustness of the discriminator significantly improved.

Perhaps surprisingly, we found that let $\pi$ maximize $\mathcal{L}_d$ will give the policy an additional corrective loss, which also leads to a valid choice of weight function $f$.
\begin{theorem}
Assume $\mathcal{L}_d(d,\log\pi)$ is twice continously differentiable with respect to $d$, and $d$ is continuously differentiable with respect to $\log\pi$. With a given discriminator $d$, then a relaxed neccessary condition for $\mathcal{L}_d(d,\log\pi)$ attains its maxima with respect to $\pi$ is to require a corrective loss term $\mathcal{L}_w$ is minimized by $\pi$, where $\mathcal{L}_w$ is given as follows:
\begin{align*}
\mathcal{L}_w = &\underset{(s, a) \sim \mathcal{D}_{e}}{\mathbb{E}} \left[\log \pi(a|s) \cdot \left(\frac{\eta}{d}+\frac{\eta}{1-d}\right) \right] \notag\\
& - \underset{(s, a) \sim \mathcal{D}_{o}}{\mathbb{E}} \left[\log \pi(a|s) \cdot \frac{1}{1-d} \right] 
% \notag\\ & + \underset{(s, a) \sim \mathcal{D}_{e}}{\mathbb{E}} \left[\log \pi(a|s) \cdot \frac{\eta}{1-d} \right]
\end{align*}
\end{theorem}
\begin{proof}
See the Proposition~\ref{app_prop} and Corollary~\ref{app_coro} in the Appendix for detailed derivation and proof.
\end{proof}

Adding the loss term $\mathcal{L}_w$ to BC task, we get the following new learning objective of $\pi$ as:
\begin{align}
\min_{\pi} \ &{\color{red} \alpha} \underset{(s, a) \sim \mathcal{D}_{e}}{\mathbb{E}} \left[-\log \pi(a|s) \right] \notag  \\ & - \underset{(s, a) \sim \mathcal{D}_{e}}{\mathbb{E}} \left[-\log \pi(a|s) \cdot \frac{\eta}{d\left(1-d\right)} \right] \notag 
\\ & +  \underset{(s, a) \sim \mathcal{D}_{o}}{\mathbb{E}} \left[-\log \pi(a|s) \cdot \frac{1}{1-d} \right], \label{eq_new_task_bc}
\end{align}
where $\alpha$ is the weight factor ($\alpha \geq 1$). 
This new objective essentially transforms the original BC task into a cost-sensitive learning problem~\citep{ling2008cost} by imposing the following weight on imitating each state-action transition as
\begin{equation}
\text {BC weights}= \begin{cases} \alpha-\eta / d(1-d), \quad & (s,a) \in \mathcal{D}_e \\ 1 /\left(1-d\right), \quad & (s,a) \in \mathcal{D}_o \end{cases},
\end{equation}
where $d$ is clipped to the range of $[0.1, 0.9]$ to satisfy the continuity assumption (see Appendix \ref{app:derivation} for details).
% Note that the derivation of the above behavioral cloning weights requires uniform continuity to be satisfied in $F$ and its derivative (details see Appendix~\ref{app:derivation}). The involvement of $1/d$ and $1/(1-d)$ may violate the continuity assumption. We thus clip the value $d$ to the range of $[0.1, 0.9]$.

Above behavioral cloning weights induce different behaviors on the imitation of transitions from $\mathcal{D}_e$ and $\mathcal{D}_o$. 
Suppose $d$ is learning in a virtuous cycle, i.e., assigns large values (close to 1) to expert transitions and small values to non-expert transitions (close to 0). The weight of those expert transitions in $\mathcal{D}_o$ will become large while the weight of those non-expert transitions will become small. For transitions in $\mathcal{D}_e$, their weights can be adjusted by tuning the parameter $\alpha$. 
Note that even if the discriminator is learned to be totally wrong (i.e., assign small values to expert transitions and large values to non-expert transitions), which may occur at the very beginning of training, the behavior cloning weights $\alpha-\eta/d(1-d)$ ($\alpha \geq 1, \eta < 1$) will not be drastically changed under value clipping. 
% still remain unchanged compared with the optimal discriminator case. 
This means that the policy can still learn from the expert dataset $\mathcal{D}_e$. Even though the weight for $\mathcal{D}_e$ is temporarily incorrect, it will be corrected as the discriminator becomes better and better.

Eq.~(\ref{eq_new_task_bc}) implies that our approach is also a variant of generalized BC objective, but uses a different form of weights.
Unlike Offline Reinforced Imitation Learning~\citep{zolna2020offline}, which uses the discriminator as the reward and learns a value function as the weight, our approach uses the discriminator outputs directly as the weight. This can greatly reduce the training time and avoid the overestimation issue in estimating the value function offline~\citep{kumar2019stabilizing}.
We term our algorithm Discriminator-Weighted Behavioral Cloning (DWBC). 
The pseudocode and implementation details of our algorithm are included in Appendix \ref{app:training}.

\section{Related Work}
\subsection{Offline Imitation Learning}
Offline IL, which has not received considerable attention, is a promising area because it makes IL more practical to satisfy critical safety desiderata. 
Offline IL methods can be folded into two paradigms: Behavioral Cloning (BC) and Offline Inverse Reinforcement Learning (Offline IRL). 

BC~\citep{pomerleau1989alvinn} is the simplest IL method that can be used in the offline setting, it considers the policy as a conditional distribution $\pi(\cdot|s)$ over actions, recent work~\citep{florence2022implicit} enhances BC by using energy-based models~\citep{lecun2006tutorial}. 
BC has shown to have no inferior performance compared to popular IL algorithms such as GAIL~\citep{ho2016generative} when clean expert demonstrations are available~\citep{ma2020adversarial}.
Unlike BC, offline IRL considers matching the state-action distribution induced by the expert policy, this can be achieved implicitly by adversarial training or explicitly by learning a reward function.
Offline IRL algorithms based on adversarial training \citep{kostrikov2019imitation,jarrett2020strictly,swamy2021moments,garg2021iq} use Intergral Probability Metrics (IPMs)~\citep{sriperumbudur2009integral} as a distance measure to solve the dual problem. They introduce a discriminator (or critic) and aim to find the saddle point of a min-max optimization problem, like GAN~\citep{goodfellow2014generative}.
% \citet{jarrett2020strictly} avoids the need of min-max problem by fixing the policy to be energy-based models, in such case the KL divergence from the demonstrator’s state-action distribution to that of the policy can be computed in closed form. 
% However, recent work finds several fundamental mathematical misconceptions in their proposed approach and we refer the reader to \citet{swamy2021moments} for more details.

The common problem of these works is that they imitate equally to all demonstrations, this will hinder the performance if $\mathcal{D}_{o}$ contain suboptimal data. To solve this, BCND reuses another policy learned by BC as the weight of original BC objective \citep{sasaki2021behavioral}, however, this requires that expert data occupies the majority of the offline dataset, otherwise the policy will be misguided to imitate the suboptimal data. ORIL first constructs a reward function that discriminates expert and exploratory trajectories, then use it to solve an offline RL problem \citep{zolna2020offline}. 
Instead of the adversarial learning scheme, the reward function can also be learned by cascading to two supervised learning steps \citep{klein2013cascaded}.

However, offline IRL based on reward learning is expensive to run, requiring solving offline RL in an inner loop, which itself is a challenging problem and prone to training instability and hyperparameter sensitivity~\citep{wu2019behavior}.
Our algorithm can be seen as a combination of these two algorithms in that
we train a discriminator to distinguish expert and non-expert data and use the output of the discriminator as the weight of the generalized BC objective, so as to imitate demonstrations selectively. 
One recent work \citep{chang2021mitigating} performs offline IL by adopting techniques from pessimistic model-based offline policy learning \citep{yu2020mopo,yu2021combo}, our work does not need to train a dynamics model nor perform the expensive min-max model-based policy optimization.
Another recent work, DemoDICE \citep{kim2021demodice}, performs offline IL with a weighted KL constraint to regularize the learned policy to stay close to both $\mathcal{D}_{e}$ and $\mathcal{D}_{o}$, it could be highly suboptimal when $\mathcal{D}_o$ contains a large collection of noisy data.
% There are also some work consider offline imitation learning from observation only.

\subsection{Offline Reinforcement Learning}
One research area highly related to offline IL is offline RL~\citep{lange2012batch,levine2020offline}, which considers performing effective RL by utilizing arbitrary given, static offline datasets, without any further environment interactions.
Note that in offline RL, the training dataset is allowed to have non-optimal trajectories and the reward for each state-action-next state transition triple is known.

Our algorithm draws connection to a branch of methods in offline RL literature that performs "filtered" behavioral cloning explicitly or implicitly. More specifically, these methods estimate an advantage function, which represents the change in expected return when taking action $a$ instead of following the current policy, and perform weighted regression based on the advantage function, defined as $\mathcal{L}_{\pi}=\mathbb{E}_{(s, a) \sim \mathcal{D}_b}\left[-\log \pi(a|s) \cdot f\left(A^{\pi}(s, a)\right) \right]$. 
The advantage $A^{\pi}$ can be estimated by Monte-Carlo methods~\citep{schulman2017proximal,peng2019advantage} or Q-value based methods~\citep{schulman2015trust,nair2020accelerating}.
The filter function $f$ can be a binary filter~\citep{wang2020critic} or an exponential filter~\citep{peng2019advantage,nair2020accelerating}.

While \citet{chen2021decision} and \citet{janner2021reinforcement} perform filtered behavioral cloning more implicitly. They cast offline RL as a sequence modeling problem and use Transformer architecture~\citep{vaswani2017attention} to perform credit assignment directly via self-attention mechanism. 
Owing to the memorization power of Transformer in capturing long-term dependencies across timesteps, these methods discard low-quality transitions and conduct behavior cloning only on high-reward transitions.

\section{Experiments}
We present empirical evaluations of DWBC in a variety of settings.
We start with describing our experimental setup, datasets and baselines. 
Then we evaluate DWBC against other baselines on a range of robotic locomotion tasks with different types of datasets.
% Finally, we analyze the property of the discriminator. Owing to the including of $\log\pi$ as input, the discriminator in DWBC can be used to perform offline policy selection~\citep{fu2020benchmarks}, which is of independent interest.
Finally, we analyze the property of the discriminator, i.e., using the discriminator to do offline policy selection~\citep{fu2020benchmarks} owing to the including of $\log\pi$ as input.

\subsection{Settings}
We construct experiments on both widely-used D4RL MuJoCo datasets~\citep{fu2020d4rl} and more complex Adroit human datasets~\citep{rajeswaranlearning}.
To verify the effectiveness of our methods, we use three setting to generate $\mathcal{D}_{e}$ and $\mathcal{D}_{o}$. Note that we use ground truth reward only to perform the data split step and discard the reward informantion afterward. 
\begin{itemize}[leftmargin=*,nosep]
\item In Setting 1, we use expert and random datasets in Mujoco environments. We sample 10 trajectories from expert datasets and 1000 trajectories from random datasets. We sample first $X\%$ trajectories from those 10 expert trajectories and combine them with those 1000 random trajectories to constitute $\mathcal{D}_{o}$, we use the remaining $1-X\%$ trajectories to constitute $\mathcal{D}_{e}$.
\item In Setting 2, we use mixed datasets in Mujoco environments. We sort from high to low of all trajectories based on the total reward summed over the entire trajectory. We define a trajectory as well-performing if it is among the top 5\% of all trajectories.
We then sample every $X^{\text{th}}$ trajectory from the well-performing trajectories to constitute $\mathcal{D}_{e}$ and use the remaining trajectories in the dataset to constitute $\mathcal{D}_{o}$.
Note that with $X$ becomes larger, $\mathcal{D}_{o}$ will contain more proportion of well-performing data. 
This setting can verify whether an algorithm effectively leverages $\mathcal{D}_{o}$ as it contains multi-level data, not only expert data in setting 1.
\item In Setting 3, we use expert and cloned datasets in Adroit environments\footnote{We use cloned datasets as low-quality datasets as there doesn't exist random datasets in D4RL Adroit environments and performing BC on cloned datasets also has a near-zero normalized score.}. We sample 100 trajectories from expert datasets and 1000 trajectories from cloned datasets.
We use the same procedure to constitute $\mathcal{D}_{e}$ and $\mathcal{D}_{o}$ as in Setting 1.
\end{itemize}

% In practice, the dataset $\mathcal{D}_{e}$ would be collected by an expert and $\mathcal{D}_{o}$ is meant to be built based on the logged interactions of the agent from the past.
% Similar to \citet{zolna2020offline}. In this paper, we build these datasets based on existing offline RL benchmark datasets. Given a logged dataset consists of multiple trajectories, we extract a small subset of well-performing trajectories to constitute demonstrations $\mathcal{D}_{e}$ and use the rest to constitute demonstrations $\mathcal{D}_{o}$. 

% We use the mixed datasets in the open D4RL benchmark datasets~\citep{fu2020d4rl}. The mixed datasets are generated by recording the replay buffer of a policy trained up to the performance of a predefined threshold (usually 1/3 of the performance of an online SAC agent~\citep{haarnoja2018soft}).
% We sort from high to low of all trajectories based on the total reward summed over the entire trajectory. We define a trajectory as well-performing if it is among the top 20\% of all trajectories.
% We then sample every $X^{\text{th}}$ trajectory from the well-performing trajectories to constitute $\mathcal{D}_{e}$ and use the remaining trajectories in the dataset to constitute $\mathcal{D}_{o}$.
% Note that with $X$ becomes larger, $\mathcal{D}_{o}$ will contain more proportion of well-performing data. 

We list all datasets used in this paper and the number of trajectories and transitions in $\mathcal{D}_{e}$ and $\mathcal{D}_{o}$ in Appendix \ref{app:dataset}, different $X$ is labeled after the dataset name.

\begin{table*}[t]
\centering
\small
\caption{Results for Mujoco and Adroit datasets. Scores are undiscounted average returns of the policy at the last 10 evaluations of training, averaged over 5 random seeds. 
All values are normalized to lie between 0 and 100, where 0 corresponds to a random policy and 100 corresponds to an expert \protect\citep{fu2020d4rl}. We bold the highest value.}
\vspace{6pt}
\resizebox{0.75\linewidth}{!}{
\begin{tabular}{c|l|rrrrr|r}
\toprule
\multicolumn{1}{l}{}                                           &                           & BC-exp        & BC-all        & BCND          & ORIL          & DemoDICE      & DWBC          \\
\midrule
\multirow{12}{*}{\rotatebox[origin=c]{90}{\textbf{Setting 1}}} & hopper\_exp-rand-30       & 74.8$\pm$11.6 & 3.1$\pm$3.1 & 2.3$\pm$0.3 & 70.7$\pm$2.5 & 42.4$\pm$6.1 & \textbf{87.2}$\pm$12.3 \\
								                               & halfcheetah\_exp-rand-30  & 24.8$\pm$5.1 & 2.2$\pm$0.0 & 2.2$\pm$0.2 & 9.8$\pm$3.6 & 2.2$\pm$0.0 & \textbf{43.9}$\pm$7.2 \\
								                               & walker2d\_exp-rand-30     & 93.9$\pm$16.3 & 1.8$\pm$2.6 & 0.4$\pm$0.0 & 2.4$\pm$1.9 & 105.0$\pm$1.7 & \textbf{106.3}$\pm$3.2 \\
								                               & ant\_exp-rand-30          & 72.7$\pm$15.7 & 34.2$\pm$3.5 & 17.0$\pm$1.2 & 64.1$\pm$21.5 & 62.2$\pm$5.5 & \textbf{91.3}$\pm$11.3 \\
                                                               & hopper\_exp-rand-60       & 75.9$\pm$16.3 & 2.6$\pm$0.6 & 2.2$\pm$0.1 & 8.0$\pm$10.4 & 57.3$\pm$6.7 & \textbf{87.4}$\pm$11.2 \\
								                               & halfcheetah\_exp-rand-60  & 6.1$\pm$1.5 & 2.2$\pm$0.0 & 2.2$\pm$0.1 & 3.6$\pm$1.5 & 2.2$\pm$0.0 & \textbf{21.3}$\pm$7.1 \\
								                               & walker2d\_exp-rand-60     & 87.6$\pm$10.8 & 1.7$\pm$1.7 & -0.2$\pm$0.0 & 5.5$\pm$1.1 & \textbf{105.8}$\pm$2.6 & \textbf{105.8}$\pm$3.3 \\
								                               & ant\_exp-rand-60          & 63.1$\pm$13.0 & 36.7$\pm$7.8 & 24.1$\pm$0.8 & 69.4$\pm$12.1 & 59.7$\pm$8.2 & \textbf{83.4}$\pm$7.2 \\
								                               & hopper\_exp-rand-90       & 24.2$\pm$14.7 & 2.3$\pm$6.5 & 2.4$\pm$1.1 & 59.5$\pm$19.7 & 38.4$\pm$15.8 & \textbf{86.7}$\pm$15.1 \\
								                               & halfcheetah\_exp-rand-90  & 1.3$\pm$2.1 & 2.2$\pm$1.2 & 2.2$\pm$0.0 & 2.2$\pm$1.3 & 2.2$\pm$0.0 & \textbf{3.4}$\pm$2.4 \\
								                               & walker2d\_exp-rand-90     & 45.8$\pm$12.4 & 0.4$\pm$7.2 & 0.4$\pm$0.3 & 0.0$\pm$17.8 & 59.6$\pm$17.2 & \textbf{90.1}$\pm$22.0 \\
								                               & ant\_exp-rand-90          & 11.1$\pm$8.7 & \textbf{31.7}$\pm$6.8 & 27.4$\pm$1.8 & \textbf{31.8}$\pm$9.4 & -63.6$\pm$0.0 & 25.7$\pm$10.3 \\
\midrule
\multirow{12}{*}{\rotatebox[origin=c]{90}{\textbf{Setting 2}}} & hopper\_mixed-2           & 41.4$\pm$8.7 & 35.0$\pm$5.1 & 2.2$\pm$1.0 & 69.8$\pm$5.2 & 1.5$\pm$0.1 & \textbf{73.0}$\pm$6.6 \\
								                               & halfcheetah\_mixed-2      & 31.9$\pm$1.8 & 36.2$\pm$0.6 & 2.2$\pm$0.8 & 36.2$\pm$4.2 & 2.2$\pm$0.3 & \textbf{38.9}$\pm$1.5 \\
								                               & walker2d\_mixed-2         & 53.2$\pm$16.3 & 32.6$\pm$2.1 & 0.3$\pm$0.0 & \textbf{65.3}$\pm$6.7 & 0.2$\pm$0.0 & 59.8$\pm$7.5 \\
								                               & ant\_mixed-2              & 70.4$\pm$5.5 & 76.1$\pm$3.5 & 29.3$\pm$2.2 & 76.6$\pm$3.1 & 30.3$\pm$2.1 & \textbf{79.1}$\pm$2.5 \\
                                                               & hopper\_mixed-5           & \textbf{68.4}$\pm$3.1 & 21.8$\pm$2.0 & 2.2$\pm$0.9 & 59.0$\pm$5.0 & 1.5$\pm$0.2 & 66.4$\pm$11.9 \\
 								                               & halfcheetah\_mixed-5      & 25.2$\pm$1.5 & 36.8$\pm$1.2 & 2.2$\pm$0.7 & \textbf{31.3}$\pm$1.9 & 2.2$\pm$0.4 & 29.0$\pm$2.7 \\
  								                               & walker2d\_mixed-5         & 48.9$\pm$3.3 & 45.7$\pm$2.0 & 0.3$\pm$0.0 & \textbf{61.3}$\pm$5.8 & 0.2$\pm$0.0 & 50.7$\pm$8.2 \\
  								                               & ant\_mixed-5              & 64.3$\pm$6.7 & 69.4$\pm$3.5 & 29.3$\pm$2.3 & 75.4$\pm$1.9 & 30.4$\pm$2.1 & \textbf{77.0}$\pm$2.9 \\
								                               & hopper\_mixed-10          & 55.2$\pm$12.3 & 16.3$\pm$6.0 & 2.2$\pm$0.2 & 59.1$\pm$4.2 & 1.5$\pm$0.2 & \textbf{61.0}$\pm$5.8 \\
								                               & halfcheetah\_mixed-10     & 14.9$\pm$0.9 & 33.0$\pm$0.6 & 2.2$\pm$0.6 & 26.1$\pm$1.9 & 2.2$\pm$0.6 & \textbf{27.8}$\pm$2.6 \\
								                               & walker2d\_mixed-10        & 25.9$\pm$0.3 & 43.9$\pm$5.3 & 0.3$\pm$0.0 & \textbf{56.4}$\pm$2.4 & 0.2$\pm$0.0 & 49.5$\pm$6.9 \\
								                               & ant\_mixed-10             & 53.4$\pm$3.8 & 68.3$\pm$4.7 & 29.3$\pm$3.1 & \textbf{75.0}$\pm$2.1 & 30.3$\pm$1.7 & 70.5$\pm$4.2 \\
\midrule
\multirow{12}{*}{\rotatebox[origin=c]{90}{\textbf{Setting 3}}} & pen\_exp-cloned-30        & 99.5$\pm$22.8 & 13.2$\pm$15.5 & -0.3$\pm$0.0 & 43.0$\pm$20.9 & 65.5$\pm$9.0 & \textbf{100.9}$\pm$10.4 \\
								                               & door\_exp-cloned-30       & \textbf{31.0}$\pm$26.9 & 0.1$\pm$0.0 & -0.1$\pm$0.0 & 0.0$\pm$0.0 & 0.0$\pm$0.0 & 5.2$\pm$3.9 \\
								                               & hammer\_exp-cloned-30     & 83.8$\pm$25.2 & 0.2$\pm$0.0 & 0.2$\pm$0.0 & 19.9$\pm$27.8 & 4.5$\pm$7.4 & \textbf{89.4}$\pm$23.6 \\
								                               & relocate\_exp-cloned-30   & 59.8$\pm$6.4 & 0.0$\pm$0.0 & -0.1$\pm$0.0 & 9.8$\pm$10.7 & 2.9$\pm$1.6 & \textbf{62.4}$\pm$5.2 \\
                                                               & pen\_exp-cloned-60        & 90.7$\pm$21.9 & 9.4$\pm$7.8 & -0.3$\pm$0.0 & 18.4$\pm$12.1 & 71.1$\pm$10.8 & \textbf{108.2}$\pm$9.4 \\
  								                               & door\_exp-cloned-60       & 5.9$\pm$7.5 & 0.0$\pm$0.0 & -0.1$\pm$0.0 & -0.1$\pm$0.0 & 0.0$\pm$0.1 & \textbf{7.5}$\pm$6.5 \\
 								                               & hammer\_exp-cloned-60     & 62.3$\pm$18.1 & 0.2$\pm$0.0 & 0.2$\pm$0.1 & 43.4$\pm$5.6 & 0.8$\pm$0.8 & \textbf{92.9}$\pm$19.1 \\
   								                               & relocate\_exp-cloned-60   & 34.0$\pm$7.4 & 0.0$\pm$0.0 & -0.1$\pm$0.0 & 2.3$\pm$0.8 & 2.1$\pm$2.0 & \textbf{56.3}$\pm$14.1 \\
								                               & pen\_exp-cloned-90        & 40.5$\pm$8.6 & 13.2$\pm$4.5 & 0.4$\pm$0.4 & 2.9$\pm$0.6 & 32.4$\pm$7.2 & \textbf{93.4}$\pm$10.4 \\
								                               & door\_exp-cloned-90       & 0.3$\pm$0.2 & -0.1$\pm$0.4 & -0.1$\pm$0.1 & -0.1$\pm$0.0 & -0.1$\pm$0.0 & \textbf{1.8}$\pm$2.8 \\
								                               & hammer\_exp-cloned-90     & \textbf{18.5}$\pm$7.4 & 0.2$\pm$0.0 & 0.2$\pm$0.0 & 0.2$\pm$0.0 & 0.2$\pm$0.1 & 14.2$\pm$8.4 \\
								                               & relocate\_exp-cloned-90   & -0.1$\pm$0.3 & 0.0$\pm$0.1 & -0.1$\pm$0.01 & -0.1$\pm$0.0 & 0.0$\pm$0.0 & \textbf{1.5}$\pm$0.8 \\
% \midrule
% \multicolumn{1}{l}{}                                           & Total                     & 105.2$\pm$1.7 & 105.2$\pm$1.7 & 105.2$\pm$1.7 & 105.2$\pm$1.7 & 105.2$\pm$1.7 & 105.2$\pm$1.7 \\
\bottomrule
\end{tabular}
}
\label{table_result}
\end{table*}

\subsection{Baseline and ablated algorithms}
We compare DWBC with the following baseline algorithms:

\noindent \textbf{BC-exp:} \quad 
Behavioral cloning on expert data $\mathcal{D}_{e}$. $\mathcal{D}_e$ owns higher quality data but with less quantity, and thus causes serious compounding error problems to the resulting policy.

\noindent \textbf{BC-all:} \quad 
Behavioral cloning on all data $\mathcal{D}_{o}$. BC-all can generalize better than BC-pos due to access to a much larger dataset, but its performance may be negatively impacted by the low-quality data in $\mathcal{D}_{o}$.

\noindent \textbf{BCND:} \quad 
% Behavioral Cloning from Noisy Demonstrations~\citep{sasaki2021behavioral} on all data. 
BCND is trained on all data, it reuses another policy learned by BC as the weight of BC, its performance will be worse if the suboptimal data occupies the major part of the offline dataset.

\noindent \textbf{ORIL:} \quad
ORIL learns a reward function and uses it to solve an offline RL problem. It suffers from large computational costs and the difficulty of performing offline RL under distributional shift. 

\noindent \textbf{DemoDICE:} \quad
The learning objective of DemoDICE contains two KL constraints, $D_{\mathrm{KL}}\left(d^{\pi} \| d^{e}\right)$ to imitate the expert and $\alpha D_{\mathrm{KL}}\left(d^{\pi} \| d^{o}\right)$ to provide proper policy regularization. 
% \noindent \textbf{DWBC-old-d:} \quad
% We include one ablation of DWBC that trains $d$ without $\log \pi$ as input, with all others remaining the same. In other words, DWBC-old-d performs new BC Task but old Discriminating Task. This ablation is to understand whether adding $\log\pi$ can make $d$ learn better.

\subsection{Comparative Evaluations}\label{sec:comp_eval}
We show the comparative results in Table~\ref{table_result} and include the learning curves in Appendix \ref{app:dataset}.
It can be shown from Table~\ref{table_result} that DWBC outperforms baseline algorithms on most tasks (27 out of 36 tasks), especially on "expert+random" datasets (setting 1\&3, 21 out of 24 tasks), showing that DWBC is well suited to make effective use of the expert dataset $\mathcal{D}_{e}$ and the mixed quality dataset $\mathcal{D}_{o}$. 

As expected, the performance of BC-exp declines as $X$ becomes larger. This is because that a larger $X$ means the number of well-performing transitions is smaller.
In some datasets (e.g., \texttt{ant\_exp-rand-90} and most in setting 2), there is no clear winner between BC-exp and BC-all, which suggests that the quality of $\mathcal{D}_o$ for the considered tasks varies. 
BCND performs poorly compared to other methods due to the majority of low-quality data in $\mathcal{D}_{o}$. It usually scores below BC-all. 
ORIL performs well in setting 2 but struggles to learn in setting 1\&3, which implies that offline RL cannot learn well with low-quality datasets. 
We also find that the performance of ORIL tends to decrease in some tasks, this "overfitting" phenomenon also occurs in experiments of offline RL papers~\citep{wu2019behavior,kumar2019stabilizing}. This is perhaps due to limited data size and model generalization bottleneck~\citep{neyshabur2017implicit}.
It can be seen that DemoDICE performs worse in setting 2 and \texttt{-90} datasets in setting 1\&3, this is because the KL regularization that regularizes the learned policy to stay close to $\mathcal{D}_{o}$ in DemoDICE is too conservative, resulting a suboptimal policy especially when $\mathcal{D}_{o}$ contains a large collection of noisy data.

% We also find that DWBC-old-d performs worse than DWBC. DWBC improve DWBC-old-d by a large margin especially when $\mathcal{D}_o$ contains more expert data (\texttt{mixed-10} datasets and \texttt{exp-rand-6} datasets), under which circumstance it is harder for the discriminator to distinguish between expert and non-expert data, without the help of $\log\pi(a|s)$.

% \begin{figure*} [t]
% \centering
% \includegraphics[width=0.45\columnwidth]{d4rl_results_legend.pdf}
% \includegraphics[width=1.0\columnwidth]{d4rl_results_new.pdf}
% \vspace{-8pt}
% \caption{Learning curves comparing the performance of DWBC against other baseline algorithms in different datasets. Curves are averaged over 5 seeds, with the shaded area representing the standard deviation across seeds. DWBC achieves the highest results in almost every task.}
% \label{fig_res}
% \vspace{-5pt}
% \end{figure*}

\begin{figure*} [t]
\centering
\subfigure {
	\includegraphics[width=0.6\columnwidth]{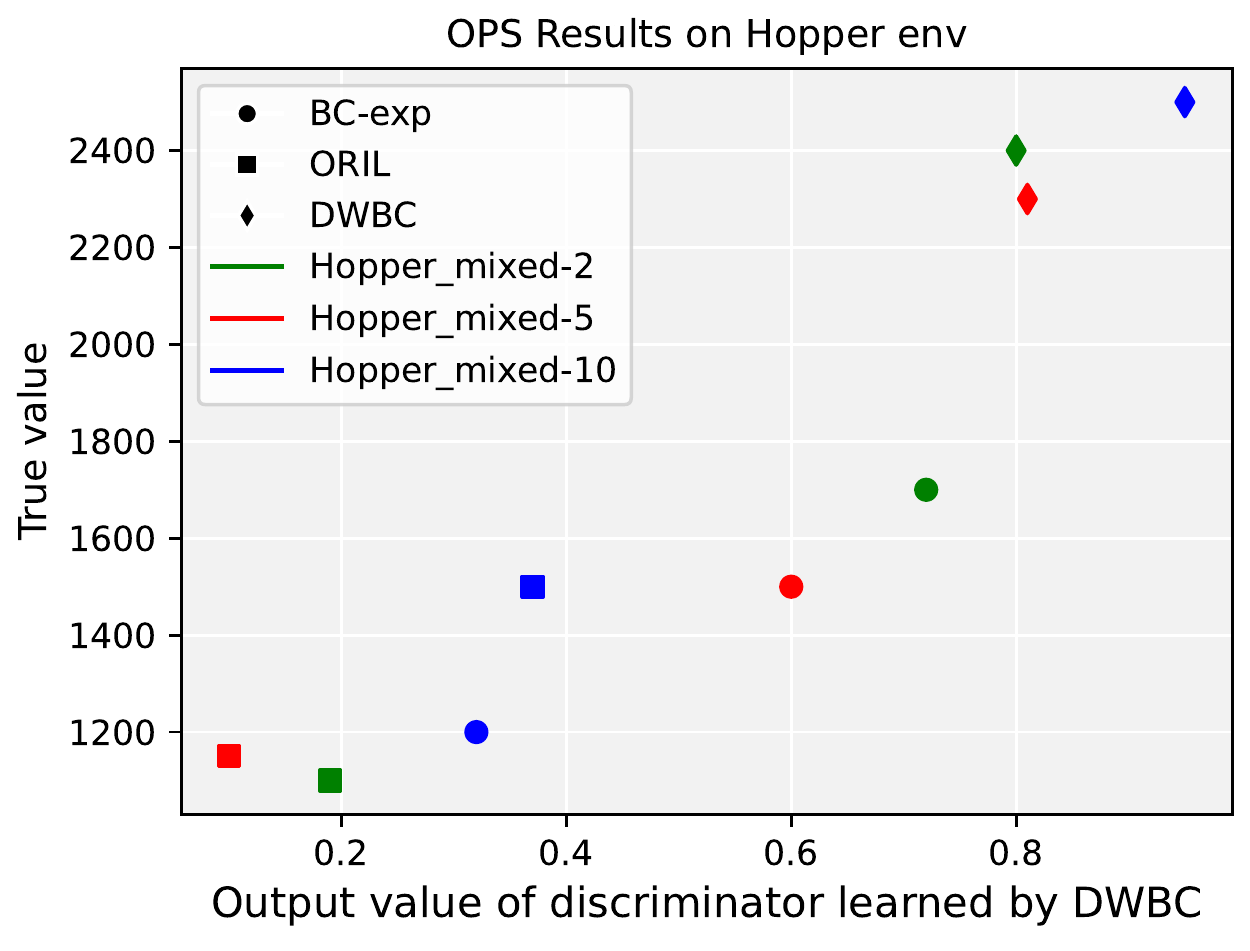} 
}
\subfigure {
	\includegraphics[width=0.6\columnwidth]{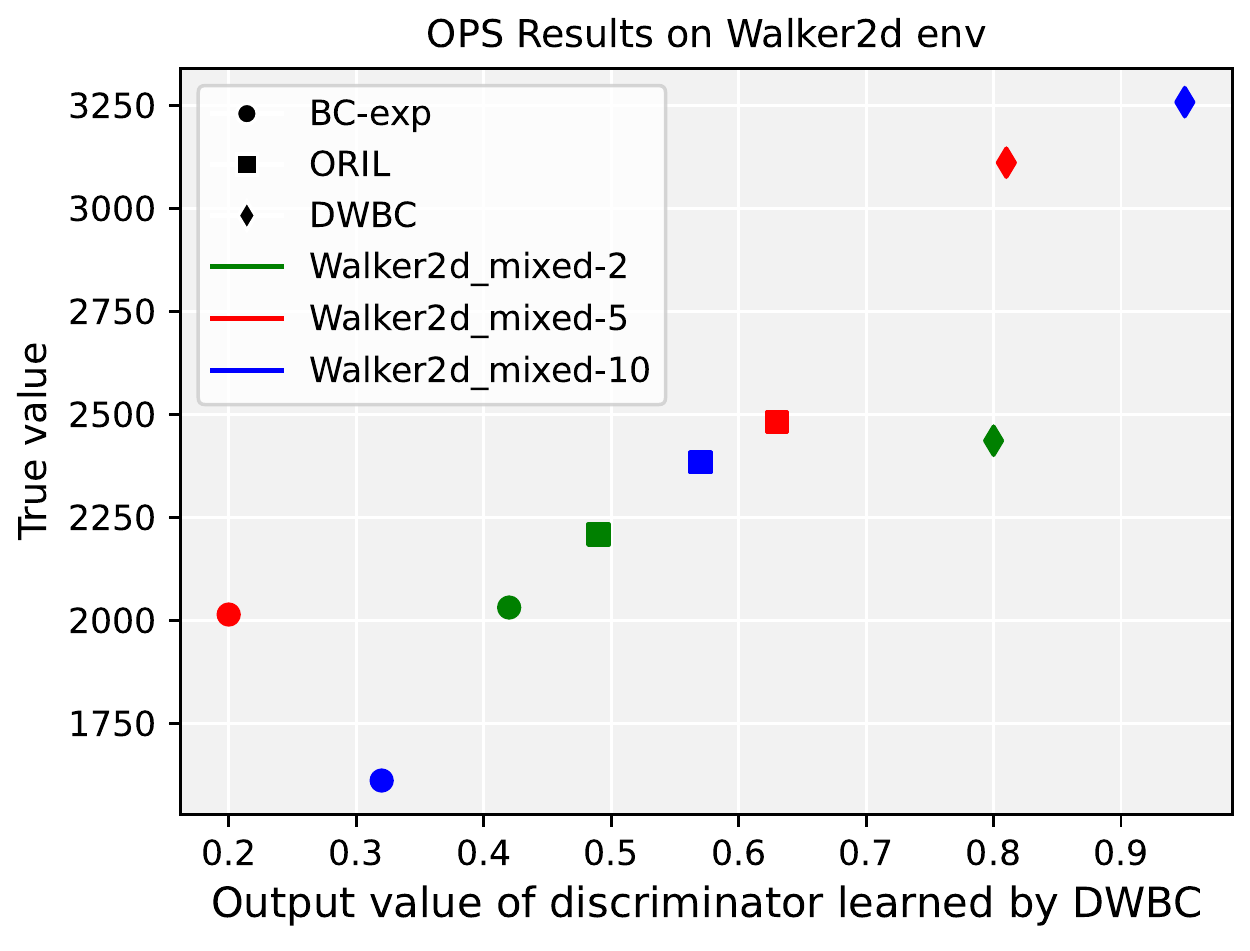} 
}
\subfigure {
	\includegraphics[width=0.6\columnwidth]{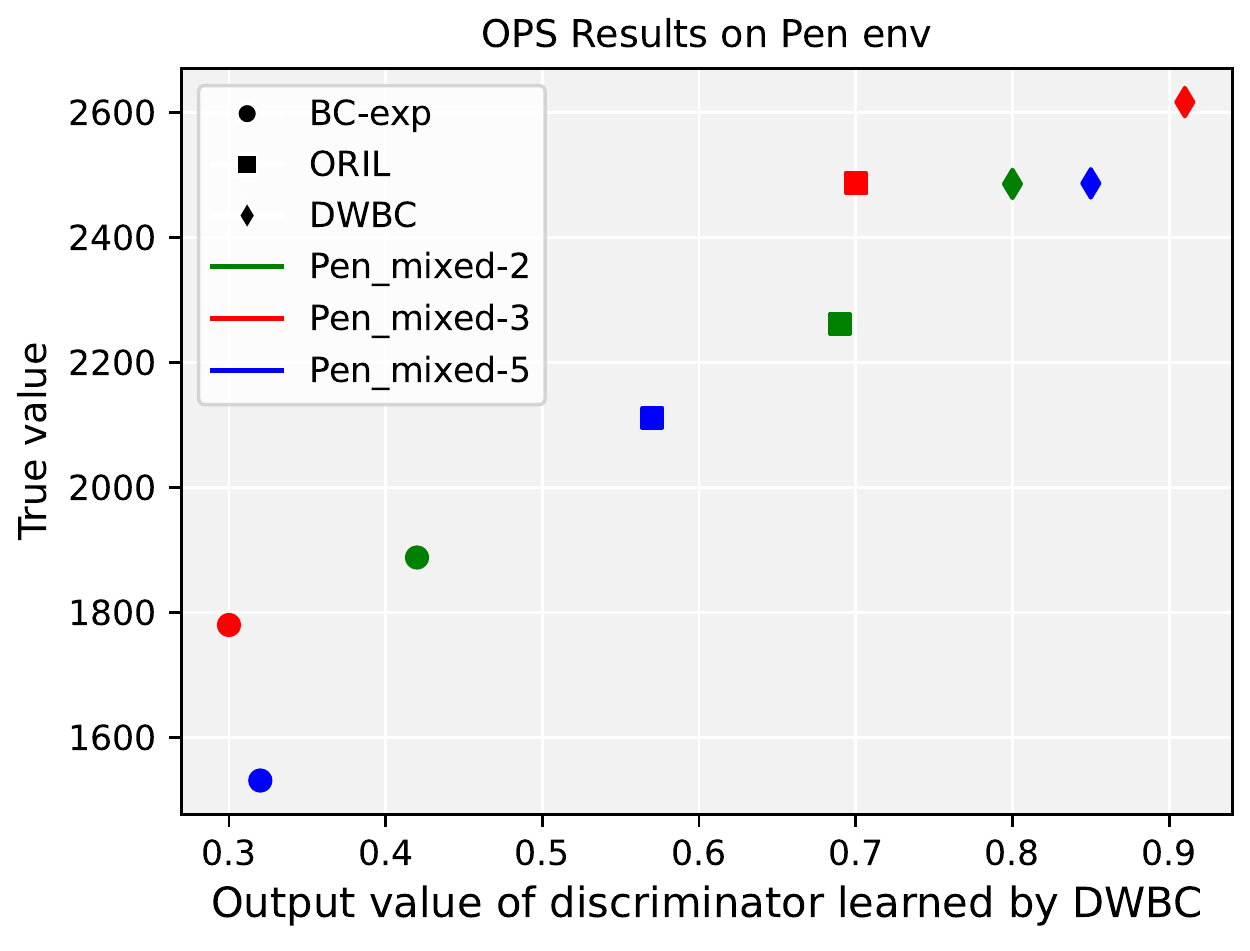} 
}
% \vspace{-5pt}
\caption{Additional experiment on offline policy selection by the discriminator learned by DWBC.}
\label{fig_ops}
\end{figure*}

\subsection{Additional Experiments}\label{sec:ops}

\noindent \textbf{Offline policy selection by the discriminator.} \quad 
Offline policy selection  (OPS)~\citep{paine2020hyperparameter,yang2020offline,dereventsov2021offline} considers the problem of choosing the best policy from a set of policies given only offline data. This problem is critical in the offline settings (i.e., offline RL and offline IL) because the online execution is often very costly and safety-aware, deploying a problematic policy may damage the real-world systems~\citep{tang2021model}.
Note that existing offline RL/IL methods break the offline assumption by evaluating different policies corresponding to their rewards in online environment interactions. However, this online evaluation is often infeasible and hence undermines the initial assumption of offline RL/IL.

We find that as a by-product, involving $\log\pi$ in the discriminator $d$ in DWBC brings an appealing characteristic, i.e., \textit{value generalization among policies}.
More specifically, $d$ values of known policies can be generalized to unknown policies, we can use expert state-action pairs from $\mathcal{D}_e$ and different policy $\pi$ as input. The discriminator will assign large values (close to 1) when the evaluated policy is close to the expert policy learned by DWBC, which also means that the evaluated policy is close to the optimal.

To validate our proposed idea, we conduct experiments in \texttt{Hopper}, \texttt{Walker2d} and \texttt{Pen} environment. In \texttt{Hopper} and \texttt{Walker2d}, we use \texttt{mixed-2}, \texttt{mixed-5} and \texttt{mixed-10} datasets, in \texttt{Pen}, we use \texttt{mixed-2}, \texttt{3} and \texttt{mixed-5} datasets. We compare three algorithms (BC-exp, ORIL and DWBC) trained in these datasets, total of 9 policies in each environment. We first train DWBC, then we use the learned discriminator $d$ along with $\mathcal{D}_e$ to compute the value $d(s,a,\log\pi_i(a|s))$ of each policy $\pi_i$.
We plot average $d(s,a,\log\pi_i(a|s))$ versus the policy's true return in Figure~\ref{fig_ops}. As shown, $d$ values well reflect the rank between almost every two policies. This means that we can first train a DWBC policy and then use the trained discriminator $d$ to do OPS, i.e., select the best policy among given candidate policies, without executing them in the environment to get the actual returns.

\begin{figure}[t]
\centering
\includegraphics[width=0.7\columnwidth]{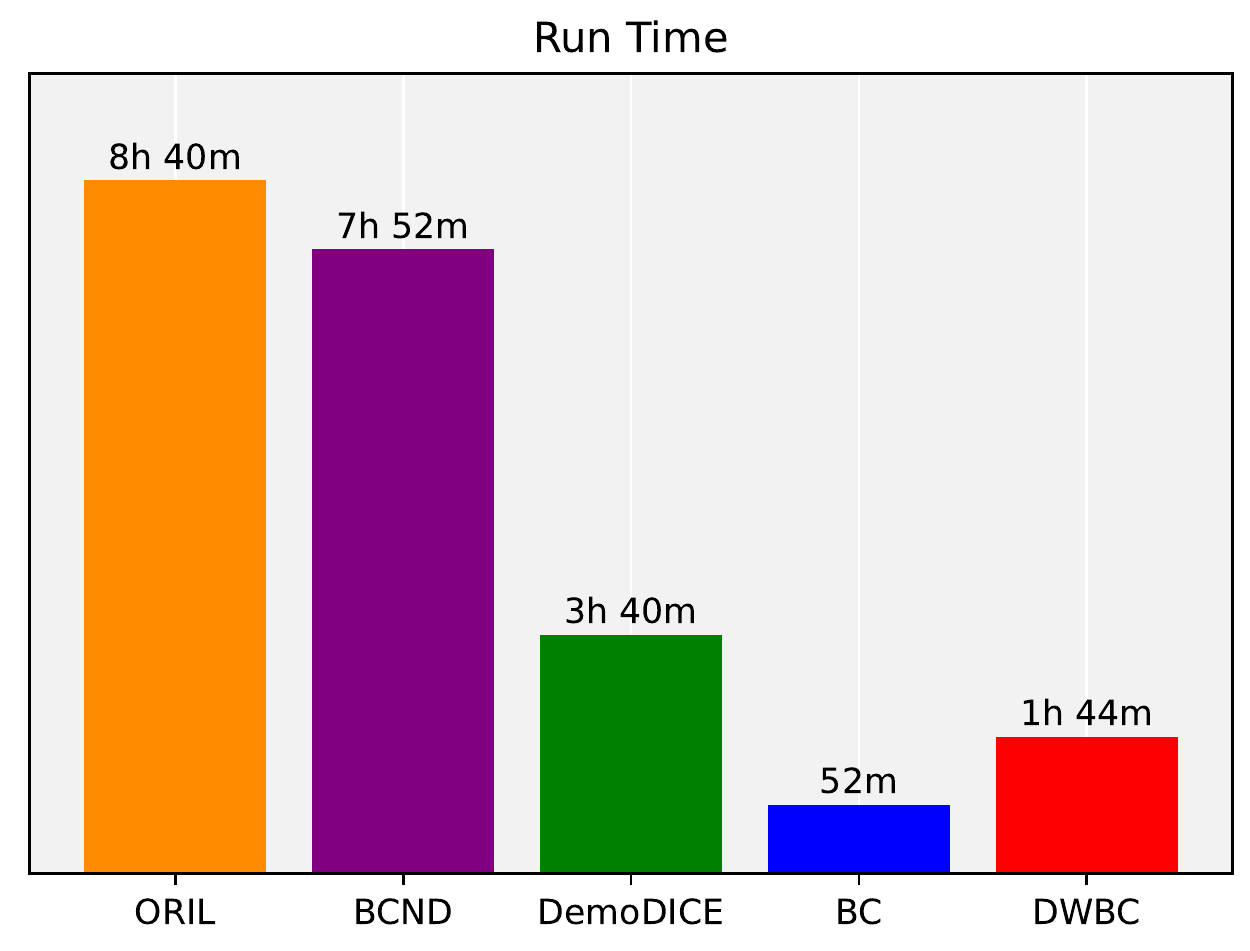}
% \vspace{-5pt}
\caption{Run time comparison of each offline IL algorithm.}
\label{fig_time}
\end{figure}

\noindent \textbf{Comparision of run time.} \quad 
We also evaluate the run time of training DWBC and other baseline algorithms for 500,000 training steps (does not include evaluation time cost). All run time experiments were executed on NVIDIA V100 GPUs. 
For a fair comparison, we use the same policy network size in BC, BCND, ORIL, DemoDICE and DWBC. The discriminator network size is also kept the same in ORIL and DWBC.
The results are reported in Figure~\ref{fig_time}. Unsurprisingly, we find the run time of our approach is only slightly more than BC, while other baselines (ORIL, BCND) are over 7 times more costly than BC.
The reason that ORIL is costly to run is due to the additional effort to solve an offline RL problem. The high computation cost of BCND is due to its inner iterations of training $K$ policy ensembles ($K=5$ in our experiment), which is also mentioned in their paper~\citep{sasaki2021behavioral}.
This demonstrates the effectiveness of DWBC by only adding a limited cost to the original BC algorithm while providing substantially improved performance.

\section{Conclusion and Future Work}
In this paper, we propose an effective and light-weighted offline imitation learning algorithm that can learn from suboptimal demonstrations without environment interactions or expert annotations. Experimental results show that our algorithm achieves higher returns and faster training speed compared to baseline algorithms, under different scenarios.
% One limitation of our work is that since our algorithm is based on weighted BC, the covariate shift problem of BC~\citep{ross2011reduction} will be inherited. That is, there is no way for the policy to learn how to recover if it deviates from the behavior policy to a new state not seen in the demonstrations.
One future work is to derive new algorithms for online IL based on our proposed cooperation framework, as recent studies~\citep{wang2021learning,eysenbach2021replacing} also reveal the importance of weighting imperfect expert demonstrations in the online IL setting.
Another future work is to consider modifying the main task from action matching to state-action distribution matching, which is known to be more robust to distributional shift~\citep{kostrikov2019imitation}.

\section*{Acknowledgements}
A preliminary version of this work was accepted on Deep RL workshop at NeurIPS 2021. 
% This work was supported by xxx.
We thank anonymous reviewers for feedback on previous versions of this paper.
This work is also supported by gifts from Haomo.AI.
% ``Data-Driven Decision Making Under Complex Autonomous Driving Scenarios'' (20222000119).

% \textbf{Do not} include acknowledgements in the initial version of
% the paper submitted for blind review.

% If a paper is accepted, the final camera-ready version can (and
% probably should) include acknowledgements. In this case, please
% place such acknowledgements in an unnumbered section at the
% end of the paper. Typically, this will include thanks to reviewers
% who gave useful comments, to colleagues who contributed to the ideas,
% and to funding agencies and corporate sponsors that provided financial

\bibliography{example_paper}
\bibliographystyle{icml2022}

\clearpage
\appendix
\onecolumn

\section{Training procedure details}
\label{app:training}
\subsection{Algorithm details}
In this section, we present the pseudocode of DWBC in Algorithm \ref{alg:dwbc}.
\begin{algorithm}
%   \small
\caption{Discriminator-Weighted Behavior Cloning (DWBC)}\label{alg:dwbc}
\begin{algorithmic}[1]
\REQUIRE Dataset $D_e$ and $D_o$, hyperparameter $\eta, \alpha$
\STATE Initialize the imitation policy $\pi$ and the discriminator $d$
\WHILE{training}
\STATE Sample $(s_e, a_e)\sim D_e$ and $(s_o, a_o)\sim D_o$ to form a training batch $\mathcal{B}$
\STATE Compute $\log\pi(a|s)$ values for samples in $\mathcal{B}$ using the learned policy $\pi$
\STATE Compute discriminator output values $d(s,a,\log\pi(a|s))$ using sampled $(s, a)$ and computed $\log\pi(a|s)$
%   $(s,a)$ from $\mathcal{B}$ and $\log\pi(a|s)$ computed from previous step
%   information $g(\tilde{Y}, \hat{Y})$ from the main task as $loss_A(\tilde{Y}, \hat{Y})$
\STATE Update $d$ by minimizing the learning objective $\mathcal{L}_d$ in Eq.(\ref{eq:loss_d}) every 100 training steps
%   \quad\quad\,\,\textbf{// Update $d$ with information from main task}
\STATE Update $\pi$ by minimizing the learning objective $\mathcal{L}_\pi$ in Eq.(\ref{eq:loss_p}) every 1 training step
%   \quad\quad\textbf{// Update $f$ with information from companion task}
\ENDWHILE
\end{algorithmic}  
\end{algorithm}

\subsection{Implementation Details}
In this paper, all experiments are implemented with Tensorflow and executed on NVIDIA V100 GPUs.
For all function approximators, we use fully connected neural networks with RELU activations. For policy networks, we use tanh (Gaussian) on outputs. 
We use Adam for all optimizers. The batch size is 256 and $\gamma$ is 0.99. 
We train each algorithm for 0.5M steps and evaluate every 5k steps. We report the average performance of the last 10 evaluations to capture each algorithm's asymptotic performance.
% We rescale the reward to $[0,1]$ as $r^{\prime}=\left(r-r_{\min }\right) /\left(r_{\max }-r_{\min }\right)$, where $r_{\max}$ and $r_{\min}$ is the maximum and the minimum reward in the dataset. Note that any affine transformation of the reward function does not change the optimal policy of the MDP. 

\begin{figure*}[h]
\centering
\includegraphics[width=0.45\columnwidth]{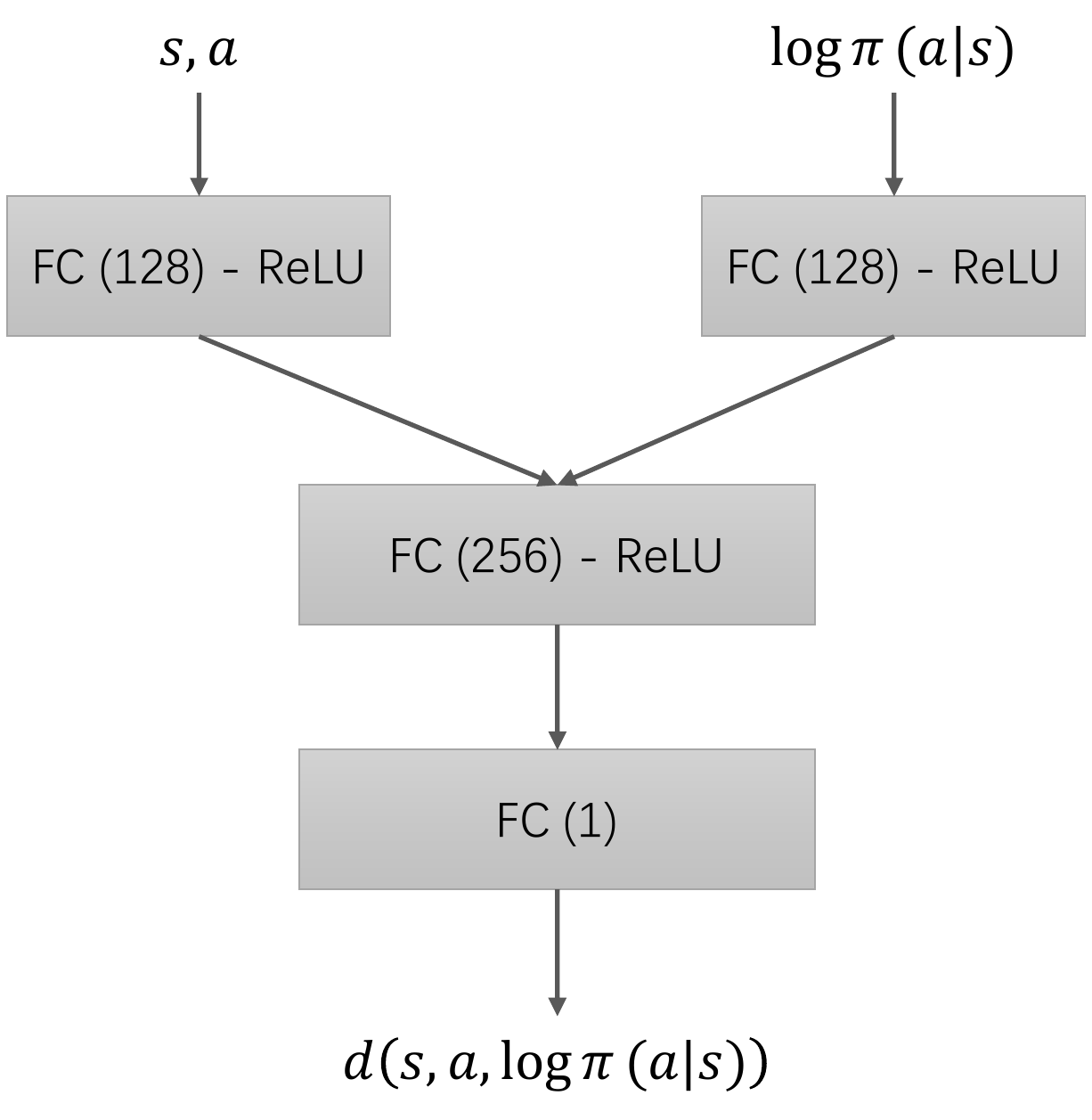} 
\caption{Structure of the discriminator network $d$.}
\label{fig_d}
\end{figure*}

For DWBC, the policy network is 2-layer MLP with 256 hidden units in each layer, 
the structure of our discriminator differs at the first hidden layer which has two input streams and each of them has 128 units, as illustrated in Figure \ref{fig_d}. We also normalize the value of $\log \pi$ to $[-20, 10]$ for better learning.
The learning rate is $1e-4$, for both the policy and the discriminator.
We use weight decay 0.005 for the policy network to prevent overfitting.
% , in each training iteration, the density ratio network is updated for 3 steps (which we find better than only 1 step). 
We clip the output of $d$ to $[0.1, 0.9]$.
We set $\alpha$ to 7.5 and $\eta$ to 0.5 for all datasets. 
% We use PU-learning objective only in setting 1, we omit it in setting 2 and 3 as we find it will do harm to the performance as the quality of random (cloned) datasets is too low.
% When using PU-learning, we set $\eta$ to 0.5.
For baselines, we run DemoDICE using the author-provided code\footnote{ \url{https://github.com/geon-hyeong/imitation-dice}}, we reimplement ORIL and BCND as we didn't find the public code.

\section{Derivation Details}
In this section, we provide the detailed design intuition and theorectical derivation of DWBC. 

\subsection{Decomposition and Reformulation of Learning Tasks}
As discussed in Section~\ref{sec:cooperative}, learning from both the expert dataset $D_e$ and suboptimal dataset $D_o$ implies the need of jointly solving two tasks: BC Task and Discriminating Task. A straightfoward solution is to learn the two tasks separately, which solves BC task by imitating the expert demonstrations in $D_e$ and learn the discriminator via PU learning using data from both $D_e$ and $D_o$:
\begin{equation*}
\begin{split}
    \text{BC Task:}\quad& \pi(a|s)\leftarrow\arg\min_{\pi} \mathcal{L}_{BC} \\
    \text{Discriminating Task:}\quad& d(s,a)\leftarrow\arg\min_{d} \mathcal{L}_{d}
\end{split}
\end{equation*}
where $\mathcal{L}_{BC}$ and $\mathcal{L}_{d}$ are discussed and given in objectives (\ref{eq_task_bc}) and (\ref{eq_task_d_pu}) in the main article as follows:
\begin{equation*}
\begin{split}
&\mathcal{L}_{BC} =\underset{(s, a) \sim \mathcal{D}_{e}}{\mathbb{E}} \left[-\log \pi(a|s) \right] \\
\mathcal{L}_{d} = \eta \underset{(s, a) \sim \mathcal{D}_{e}}{\mathbb{E}} \left[-\log d(s,a) \right] &+ \underset{(s, a) \sim \mathcal{D}_{o}}{\mathbb{E}} \left[-\log(1-d(s,a)) \right] - \eta \underset{(s, a) \sim \mathcal{D}_{e}}{\mathbb{E}} \left[-\log(1-d(s,a)) \right]
\end{split}
\end{equation*}
Na\"ively solving above two tasks separately is insufficient. First, the BC task only learn from the expert dataset $D_e$, fails to utilize the potential valuable information in the supoptimal dataset $D_o$.
Second, as discussed in Section \ref{sec:cooperative}, both tasks lack sufficient information to improve their own performance. 
For example, a good discriminator could provide important information to distinguish the potential expert samples in the suboptimal dataset $D_e$, which are valuable for the learning of policy $\pi$; a well-performed policy $\pi(a|s)$ will assing large probabilities to expert actions under expert states, which could provide additional learning signal for the discriminator $d$ to more easily contrast expert and non-expert transitions in $D_o$. 

There are two existing approaches can be used to jointly solve above two tasks, however, both of them have some drawbacks. One approach is to cast the problem into a multi-task style multi-objective optimziation problem, by optimizing an augmented loss $\beta\mathcal{L}_{BC} + (1-\beta)\mathcal{L}_d, \beta\in(0,1)$ for both $\pi$ and $d$. The problem is that the BC Task and the Discriminating Task are different tasks, the potential contradiction of the two tasks in certain settings may impede both tasks from achieving the best perforamnce. Moreover, properly selecting the hyperparameter $\beta$ is very tricky. Another approach is to adopt a GAN-style model \citep{goodfellow2014generative} which treats the policy as the generator and optimize it implicitly through solving a min-max optimization problem with the discriminator loss $\mathcal{L}_d$. However, this is very costly and is known to suffer from training instability and issues such as mode collapse \citep{arjovsky2017wasserstein}. Moreover, although we have an explict loss function $\mathcal{L}_{BC}$ for $\pi$, it is not used in such a GAN-style model, which results in potential loss of information.

In this paper, we design an new cooperative learning mechansim to address above issues, which also results in a computationally efficient practical algorithm. It includes three key ingredients: 1) sharing information between the BC Task and the Discriminating Task to achieve cooperative learning; 2) enabling the BC Task to learn on both expert and suboptimal data by introduce an additional corrective loss $\mathcal{L}_w$ impacted by the discriminator outputs; 3) solving both tasks in fully supervised learning manner to maintain computational efficiency. In our approach, we consider following alternative formulation to establish information sharing across the two tasks and enable cooperative learning:
% (see Fig.~\ref{fig_illustration} for a more intuitive illustration):
\begin{equation}
\label{eq_reformulation}
\begin{split}
    \text{New BC Task:}\quad& \pi(a|s)\leftarrow\arg\min_{\pi} \alpha \mathcal{L}_{BC} + {\color{red}\mathcal{L}_{w}}, \quad \alpha > 1 \\
    \text{New Discriminating Task:}\quad& d(s,a, {\color{red} \log \pi(a|s)})\leftarrow\arg\min_{d} \mathcal{L}_{d}
\end{split}
\end{equation}
with the new $\mathcal{L}_d$ given in objective (\ref{eq_task_d_pu}) as follows:
\begin{equation*}
\begin{split}
\mathcal{L}_d = &\eta \underset{(s, a) \sim \mathcal{D}_{e}}{\mathbb{E}} \left[-\log d(s,a,{\color{red} \log \pi(a|s)}) \right] + \underset{(s, a) \sim \mathcal{D}_{o}}{\mathbb{E}} \left[-\log(1-d(s,a,{\color{red} \log \pi(a|s)})) \right] \\
&- \eta \underset{(s, a) \sim \mathcal{D}_{e}}{\mathbb{E}} \left[-\log(1-d(s,a,{\color{red} \log \pi(a|s)})) \right]
\end{split}
\end{equation*}

In above reformulation, we design the information provided by the policy to discriminator as the element-wise imitation loss value $\log\pi(a|s)$, and the information provided to the policy as the addional corrective loss term $\mathcal{L}_w$ computed using output values of the discriminator $d$ on samples from both $D_e$ and $D_o$ (i.e., $D_b$). With the involvement of $\mathcal{L}_w$, we can allow the policy $\pi$ learning from sub-optimal data under the guidance of the discriminator. Moreover, to more robustly learn the discriminator $d$, we borrow the idea of adversairal training \citep{lowd2005adversarial} and make the policy $\pi$ challenge $d$ by maximizing $\mathcal{L}_d$.

We will show in the next section that with the choice of element-wise imitation loss $\log\pi(a|s)$ as the form of information and the advesarial behavior of policy $\pi$, an exact form of $\mathcal{L}_w$ can be derived, and eventually, transforms the original BC task into a cost sensitive learning problem.

\subsection{Drivation of the Corrective Loss Term $\mathcal{L}_w$}
\label{app:derivation}

In this section, we resort to functional analysis and calculus of variation to derive the exact form of $\mathcal{L}_w$. Under the reformulated problem (\ref{eq_reformulation}), both the discriminator $d$ and its loss $\mathcal{L}_d$ are impacted by the information provided by policy $\pi$ ($\log\pi(a|s)$). Hence they are now functional of $\pi$ (i.e., function of a function). For simplicity, we can express functional $d$ and $\mathcal{L}_d$ as $d(s,a,\log\pi(a|s))$ and $\mathcal{L}_d(d,\log\pi)$.
We are interested to see how the variation of $\pi$ impacts $\mathcal{L}_d$, and further influence $d$. Moreover, we can use a specific form of $\mathcal{L}_w$ to alter the behavior of the learned $\pi$ to achieve the desired adversarial behavior.

By inspecting the form of $\mathcal{L}_d$, note that we can equivalently write it as following integral form:
\begin{align}
\mathcal{L}_d(d,\log\pi) = &\eta \underset{(s, a) \sim \mathcal{D}_{e}}{\mathbb{E}} \left[-\log d(s,a,\log \pi(a|s)) \right] + \underset{(s, a) \sim \mathcal{D}_{o}}{\mathbb{E}} \left[-\log(1-d(s,a,\log \pi(a|s))) \right]\notag\\
&- \eta \underset{(s, a) \sim \mathcal{D}_{e}}{\mathbb{E}} \left[-\log(1-d(s,a,\log \pi(a|s))) \right] \label{eq:loss_d}\\
=& \int_{\Omega_s}\int_{\Omega_a} \Big[P_{\mathcal{D}_{e}}(s,a)\cdot\eta \left[-\log d(s,a,\log \pi(a|s)) \right] + P_{\mathcal{D}_{o}}(s,a)\left[-\log(1-d(s,a,\log \pi(a|s))) \right] \notag\\
&- P_{\mathcal{D}_{e}}(s,a)\cdot \eta \left[-\log(1-d(s,a,\log \pi(a|s))) \right]
\Big]\text{d}s\text{d}a \notag\\
\triangleq& \int_{\Omega_s}\int_{\Omega_a} F(s,a,d,\log\pi(a|s))\text{d}s\text{d}a \label{eq:integral}
\end{align}
where $P_{\mathcal{D}_{e}}(s,a)$, $P_{\mathcal{D}_{o}}(s,a)$ is the prabability distribution for state-action pair (s,a) in dataset $\mathcal{D}_{e}$ and $\mathcal{D}_{o}$, and $\Omega_s, \Omega_a$ are the domain for state $s$ and action $a$ under $\mathcal{D}_b$. Observe that the functional $\mathcal{L}_d(d,\log\pi)$ has a natural integral form given the following functional $F(s,a,d,\log\pi(a|s))$:
\begin{equation}\label{eq:app_F}
\begin{aligned}
F(s,a,d,\log\pi(a|s)) = &P_{\mathcal{D}_{e}}(s,a)\cdot\eta \left[-\log d(s,a,\log \pi(a|s)) \right] + P_{\mathcal{D}_{o}}(s,a)\left[-\log(1-d(s,a,\log \pi(a|s))) \right] \\
&- P_{\mathcal{D}_{e}}(s,a)\cdot \eta \left[-\log(1-d(s,a,\log \pi(a|s))) \right]
\end{aligned}
\end{equation}

A functional defined in an integral form like Eq. (\ref{eq:integral}) is commonly studied in functional analysis and calculus of variation~\citep{gelfand2000calculus}.
To enforce the adversarial behavior of $\pi$, we want to make $\pi$ challenge the discriminator $d$ by maximizing $\mathcal{L}_d(d,\log\pi)$.
By doing so, the policy is finding "adversarial attacks" for $\mathcal{L}_d$ such that minimizing $\mathcal{L}_d(d,\log\pi)$ becomes harder for the discriminator. This essentially lead to the following min-max optimization problem for $\mathcal{L}_d(d,\log\pi)$, and can be seen as minimizing the worst-case error, which makes the robustness of the discriminator significantly improved \cite{carlini2019evaluating,fawzi2016robustness,goodfellow2014explaining}.
\begin{equation}\label{eq:app_minimax}
    \min_d\max_\pi \mathcal{L}_d(d,\log\pi)
\end{equation}

% Maximizing $\mathcal{L}_{d}$ for $\pi$ under current $d$ is equivalent to finding the maxima of functional $J(\pi)$ ($\mathcal{L}_d$ with $\theta_d$ and $d$ fixed). 
Directly solving above min-max optimization problem can be highly complex. To simplify the analysis, we focus on the inner maximization problem for $\pi$ and derive an necessary condition that leads to a tractable form of the corrective loss term $\mathcal{L}_w$. Consider $d$ as an unknown external functional decided by the outer minimization problem, maximizing $\mathcal{L}_d(d,\log\pi)$ with respect to $\pi$ requries to find the maxima of functional $\mathcal{L}_d(d,\log\pi)$.
We can show with following proposition that a relaxed condition is needed to be satisfied.

\begin{proposition}\label{app_prop}
With a given discriminator $d$ decided by the outer maximization problem of (\ref{eq:app_minimax}), and functional $F(s,a,d,\log\pi(a|s))$ defined in Eq.(\ref{eq:app_F}), if continuity of both $F$ and $d$ and its derivatives are satisfied, a relaxed neccessary condition for $\mathcal{L}_d(d,\log\pi)$ attaining its extrema with respect to $\pi$ is: 
\begin{equation}
    \int_{\Omega_s}\int_{\Omega_a} \frac{\partial F(s,a,d,\log\pi(a|s))}{\partial d(s,a,\log\pi(a|s))}\cdot \nabla_{\theta_\pi}\log\pi(a|s) \text{\textnormal{d}}s\text{\textnormal{d}}a = 0
\end{equation}
where $\theta_\pi$ is the model parameters of $\pi$.
\end{proposition}
\begin{proof}
According to the calculus of variations~\citep{gelfand2000calculus}, the extrema (maxima or minima) of functional $\mathcal{L}_d(d,\log\pi)$ with respect to $\pi$ ($d$ is a given function and considered as fixed) can be obtained by solving the associate Euler-Langrangian equation as follows:
\begin{equation*}
    F_\pi - \frac{\partial}{\partial s}F_{\frac{\partial \pi}{\partial s}} - \frac{\partial}{\partial a}F_{\frac{\partial \pi}{\partial a}} = 0
\end{equation*}
where $F_f$ represents $\frac{\partial F}{\partial f}$. In our case, $\frac{\partial \pi}{\partial s}$ and $\frac{\partial \pi}{\partial a}$ does not appear in the form of $F$, hence the later two terms are zero. Therefore, it is necessary that the following functional equation holds:
\begin{equation*}
    F_\pi = \frac{\partial F}{\partial d}\cdot\frac{\partial d}{\partial \log\pi}\cdot\frac{\partial \log\pi}{\pi} = 0
\end{equation*}
Consider $\theta_\pi$ as the model parameter of $\pi$, above equation also suggests that 
\begin{equation*}
    \frac{\partial F}{\partial d}\cdot\frac{\partial d}{\partial \log\pi}\cdot\frac{\partial \log\pi}{\pi}\cdot \frac{\partial \pi}{\partial \theta_\pi} = \frac{\partial F}{\partial d}\cdot\frac{\partial d}{\partial \log\pi}\cdot\nabla_{\theta_\pi}\log\pi =0
\end{equation*}

As both $d$ and $F$ are real-valued functions, hence the same with their derivatives $\partial F/\partial d$ and $\partial d/\partial \log\pi$. Moreover, by assumption, the continuity of $\partial F/\partial d$ and $\partial d/\partial \log\pi$ is satisfied, as the set of real-valued continuous functions is a commutative ring \cite{hewitt1948rings}, thus their order in above equation can be swapped. We have
\begin{equation}\label{eq:app_euler_final}
    \frac{\partial d}{\partial \log\pi}\cdot\frac{\partial F}{\partial d}\cdot\nabla_{\theta_\pi}\log\pi =0
\end{equation}

Note that $d$ is determined by the outer minimization problem of (\ref{eq:app_minimax}), thus $\partial d/\partial \log\pi$ is unknown and not obtainable by solely inspecting the inner maximization problem. To ensure above functional equation holds for any $(s,a)$ in $\Omega_s\times\Omega_a$, we instead consider another solution of the functional equation Eq. (\ref{eq:app_euler_final}) by letting $\frac{\partial F}{\partial d}\cdot\nabla_{\theta_\pi}\log\pi=0$. Directly solving the new equation is still intractable, since both $F$ and $\nabla_{\theta_\pi}\log\pi$ can be complicated functions, and the data distribution $P_{\mathcal{D}_e}(s,a)$ and $P_{\mathcal{D}_o}(s,a)$ is typically unknown. However, we can obtain a relaxed and tractable condition by computing the integration of  $\frac{\partial F}{\partial d}\cdot\nabla_{\theta_\pi}\log\pi$. Since both the datasets $\mathcal{D}_e$ and $\mathcal{D}_o$ are finite and fixed, the domain of $s$ and $a$, $\Omega_s$ and $\Omega_a$, are also closed and bounded, hence the final integration is still 0, that is

\begin{equation}\label{eq:app_int_final}
    \int_{\Omega_s}\int_{\Omega_a} \frac{\partial F}{\partial d}\cdot\nabla_{\theta_\pi}\log\pi \text{d}s\text{d}a =0
\end{equation}

\end{proof}

We are interested in the relaxed condition (\ref{eq:app_int_final}) because it is computational feasible and we can use it to derive the exact form of $\mathcal{L}_w$.
\begin{corollary}\label{app_coro}
The relaxed condition (\ref{eq:app_int_final}) can be satisfied by minimizing the corrective loss term $\mathcal{L}_w$ of the following form with respect to $\theta_\pi$:
\begin{equation}
\label{eq:app_addtional_loss}
\mathcal{L}_w = \underset{(s, a) \sim \mathcal{D}_{e}}{\mathbb{E}} \left[\log \pi(a|s) \cdot \frac{\eta}{d} \right] - \underset{(s, a) \sim \mathcal{D}_{o}}{\mathbb{E}} \left[\log \pi(a|s) \cdot \frac{1}{1-d} \right] + \underset{(s, a) \sim \mathcal{D}_{e}}{\mathbb{E}} \left[\log \pi(a|s) \cdot \frac{\eta}{1-d} \right]
\end{equation}
where $d$ represents $d(s,a,\log\pi(a|s))$ for simplicity. 
\end{corollary}
\begin{proof}
Observe that

\begin{equation}\label{eq:app_int_expanded}
\begin{aligned}
    0 =&\int_{\Omega_s}\int_{\Omega_a} \frac{\partial F(s,a,d,\log\pi(a|s))}{\partial d(s,a,\log\pi(a|s))}\cdot \nabla_{\theta_\pi}\log\pi(a|s) \text{d}s\text{d}a \\
    =& \int_{\Omega_s}\int_{\Omega_a} \Bigg[
    -P_{\mathcal{D}_e}(s,a)\cdot \left[\frac{\eta}{d(s,a,\log \pi(a|s))} \right] +
    P_{\mathcal{D}_o}(s,a)\left[\frac{1}{1-d(s,a,\log \pi(a|s))} \right] \\
&- P_{\mathcal{D}_e}(s,a)\cdot \left[\frac{\eta}{1-d(s,a,\log \pi(a|s))} \right]
    \Bigg] \cdot\nabla_{\theta_\pi}\log\pi(a|s) \text{d}s\text{d}a \\
    =& -\underset{(s, a) \sim \mathcal{D}_{e}}{\mathbb{E}} \left[\frac{\eta}{d} \cdot \nabla_{\theta_\pi}\log\pi(a|s) \right]
+\underset{(s, a) \sim \mathcal{D}_{o}}{\mathbb{E}} \left[\frac{1}{1-d} \cdot \nabla_{\theta_\pi}\log\pi(a|s) \right] 
- \underset{(s, a) \sim \mathcal{D}_{e}}{\mathbb{E}} \left[\frac{\eta}{1-d} \cdot \nabla_{\theta_\pi}\log\pi(a|s) \right]
\end{aligned}
\end{equation}
where in the last equation, we slightly abuse the notations and write the output value of $d(s,a,\log\pi(a|s))$ as $d$ for simplicity.
Above condition can be equivalently perceived as the first-order optimality condition of the loss term $\mathcal{L}_w$ specified in Eq. (\ref{eq:app_addtional_loss}), i.e., derivative equal to zero. 
% $\partial \mathcal{L}_w/\partial \theta_\pi$ has exactly the same form of Eq. (\ref{eq:app_int_expanded}). 

Comparing $\partial \mathcal{L}_w/\partial \theta_\pi$ and the final form of Eq. (\ref{eq:app_int_expanded}), note that we introduce a minus sign on $\mathcal{L}_w$. This is to ensure that by minimizing $\mathcal{L}_w$, we are updating in the gradient ascent direction of $\mathcal{L}_d(d,\log\pi)$ and find its maxima rather than minima. Hence minimizing $\mathcal{L}_w$ with respect to $\pi$ (make $\partial \mathcal{L}_w/\partial \theta_\pi=0$) satisfies the relaxed condition (\ref{eq:app_int_final}), which is derived from the neccessary condition of solving the inner maximization problem of $\mathcal{L}_d(d,\log\pi)$ specified in (\ref{eq:app_minimax}).
% Hence by minimizing $\mathcal{L}_w$ with respect to $theta_\pi$, we can satisfy the relaxed condition (\ref{eq:app_int_final}).
\end{proof}

Adding the new corrective loss term $\mathcal{L}_w$ back to our reformulated problem (\ref{eq_reformulation}), we obtain the final learning objective of $\pi$ for our BC task (Eq.(\ref{eq_new_task_bc}) in the main article):
\begin{equation}
\label{eq:loss_p}
\min_{\pi} {\color{red} \alpha} \underset{(s, a) \sim \mathcal{D}_{e}}{\mathbb{E}} \left[-\log \pi(a|s) \right] {\color{red} - \underset{(s, a) \sim \mathcal{D}_{e}}{\mathbb{E}} \left[-\log \pi(a|s) \cdot \frac{\eta}{d\left(1-d\right)} \right] + \underset{(s, a) \sim \mathcal{D}_{o}}{\mathbb{E}} \left[-\log \pi(a|s) \cdot \frac{1}{1-d} \right]},\quad \alpha>1
\end{equation}

Note that the derivation of $\mathcal{L}_w$ requires continuity to be satisfied in $\partial F/\partial d$. The involvement of discriminator output values $1/d(s,a,\log\pi(a|s))$ and $1/(1-d(s,a,\log\pi(a|s))$ may violate the continuity assumption. We thus clip the discriminator output values to the range of $[0.1, 0.9]$ in our practical algorithm.

\section{Additional results}
\label{app:dataset}
\subsection{Datasets details}
In Table \ref{table_dataset}, we list all datasets used in our paper and the number of trajectories and transitions in $\mathcal{D}_{e}$ and $\mathcal{D}_{o}$, where different $X$ is labeled after the dataset name.
\begin{table*}[b]
\small
\centering
\caption{Dataset details.}
\vspace{5pt}
\begin{tabular}{c|cc|cc}
\midrule
\multirow{2}{*}{\textbf{Dataset-$X$}} & \multicolumn{2}{c|}{$\bm{\#\mathcal{D}_e}$} & \multicolumn{2}{c}{$\bm{\#\mathcal{D}_o}$} \\ \cmidrule{2-5} 
                  & Trajectories          & Transitions         & Trajectories         & Transitions         \\ 
\midrule
Hopper\_exp-rand-30                     & 7   & 7,000  & 1,003 & 24,723 \\
Hopper\_exp-rand-60                     & 4   & 4,000  & 1,006 & 27,723 \\
Hopper\_exp-rand-90                     & 1   & 1,000  & 1,009 & 30,723 \\
Hopper\_mixed-2                         & 51  & 38,830 & 1,989 & 363,026 \\
Hopper\_mixed-5                         & 21  & 15,286 & 2,019 & 386,570 \\
Hopper\_mixed-10                        & 11  & 8,263  & 2,029 & 393,593 \\
\midrule
Halfcheetah\_exp-rand-30                & 7   & 7,000  & 1,003 & 1,002,999\\
Halfcheetah\_exp-rand-60                & 4   & 4,000  & 1,006 & 1,005,999\\
Halfcheetah\_exp-rand-90                & 1   & 1,000  & 1,009 & 1,008,999\\
Halfcheetah\_mixed-2                    & 5   & 5,000  & 196   & 196,000 \\
Halfcheetah\_mixed-5                    & 2   & 2,000  & 199   & 199,000 \\
Halfcheetah\_mixed-10                   & 1   & 1,000  & 200   & 200,000\\
\midrule
Walker2d\_exp-rand-30                   & 7   & 7,000  & 1,003 & 22,877\\
Walker2d\_exp-rand-60                   & 4   & 4,000  & 1,006 & 25,877\\
Walker2d\_exp-rand-90                   & 1   & 1,000  & 1,009 & 28,877\\
Walker2d\_mixed-2                       & 27  & 26,375 & 1,065 & 274,625\\
Walker2d\_mixed-5                       & 11  & 10,789 & 1,081 & 290,211\\
Walker2d\_mixed-10                      & 6   & 5,820  & 1,086 & 295,180\\
\midrule
Ant\_exp-rand-30                        & 7   & 6,465  & 1,003 & 183,912 \\
Ant\_exp-rand-60                        & 4   & 4,000  & 1,006 & 186,377 \\
Ant\_exp-rand-90                        & 1   & 1,000  & 1,009 & 189,377 \\
Ant\_mixed-2                            & 12  & 11,876 & 472   & 289,124 \\
Ant\_mixed-5                            & 5   & 5,000  & 479   & 296,000 \\
Ant\_mixed-10                           & 3   & 3,000  & 481   & 298,000 \\
\midrule
Pen\_exp-cloned-30                      & 70   & 7,000    & 1,030    & 102,862 \\
Pen\_exp-cloned-60                      & 40   & 4,000    & 1,060    & 105,862 \\
Pen\_exp-cloned-90                      & 10   & 1,000    & 1,090    & 108,862 \\
\midrule
Door\_exp-cloned-30                     & 70   & 14,000   & 1,030    & 206,000 \\
Door\_exp-cloned-60                     & 40   & 8,000    & 1,060    & 212,000 \\
Door\_exp-cloned-90                     & 10   & 2,000    & 1,090    & 218,000 \\
\midrule
Hammer\_exp-cloned-30                   & 70   & 14,000   & 1,030    & 206,000 \\
Hammer\_exp-cloned-60                   & 40   & 8,000    & 1,060    & 212,000 \\
Hammer\_exp-cloned-90                   & 10   & 2,000    & 1,090    & 218,000 \\
\midrule
Relocate\_exp-cloned-30                 & 70   & 14,000   & 1,030    & 206,000 \\
Relocate\_exp-cloned-60                 & 40   & 8,000    & 1,060    & 212,000 \\
Relocate\_exp-cloned-90                 & 10   & 2,000    & 1,090    & 218,000 \\
\midrule
\end{tabular}
\label{table_dataset}
\end{table*}

\subsection{Learning curves}
\begin{figure*}[b]
\centering
\includegraphics[width=0.9\columnwidth]{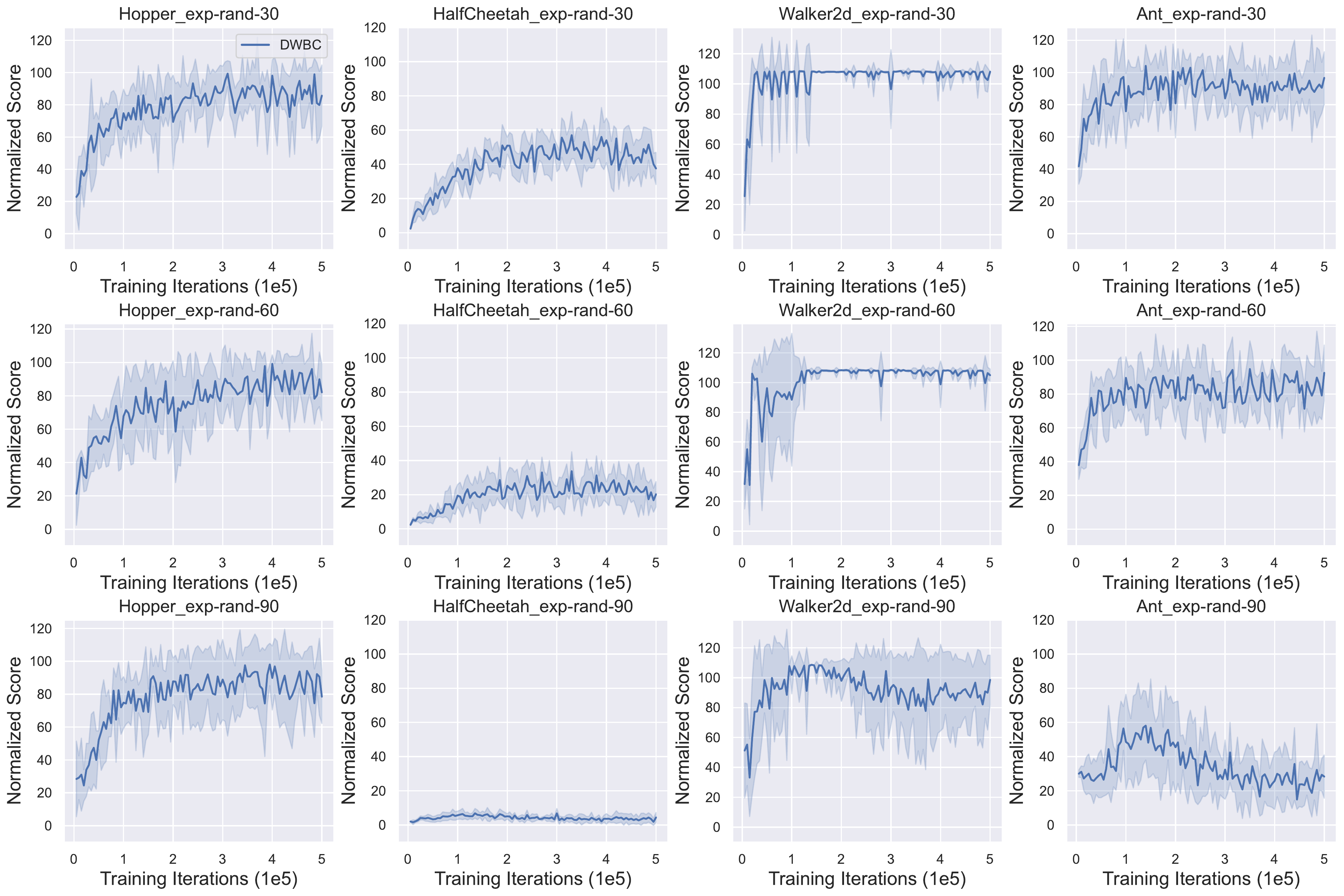} 
\caption{Learning curves for Setting 1.}
\label{fig_result1}
\end{figure*}

\begin{figure*}[b]
\centering
\includegraphics[width=0.9\columnwidth]{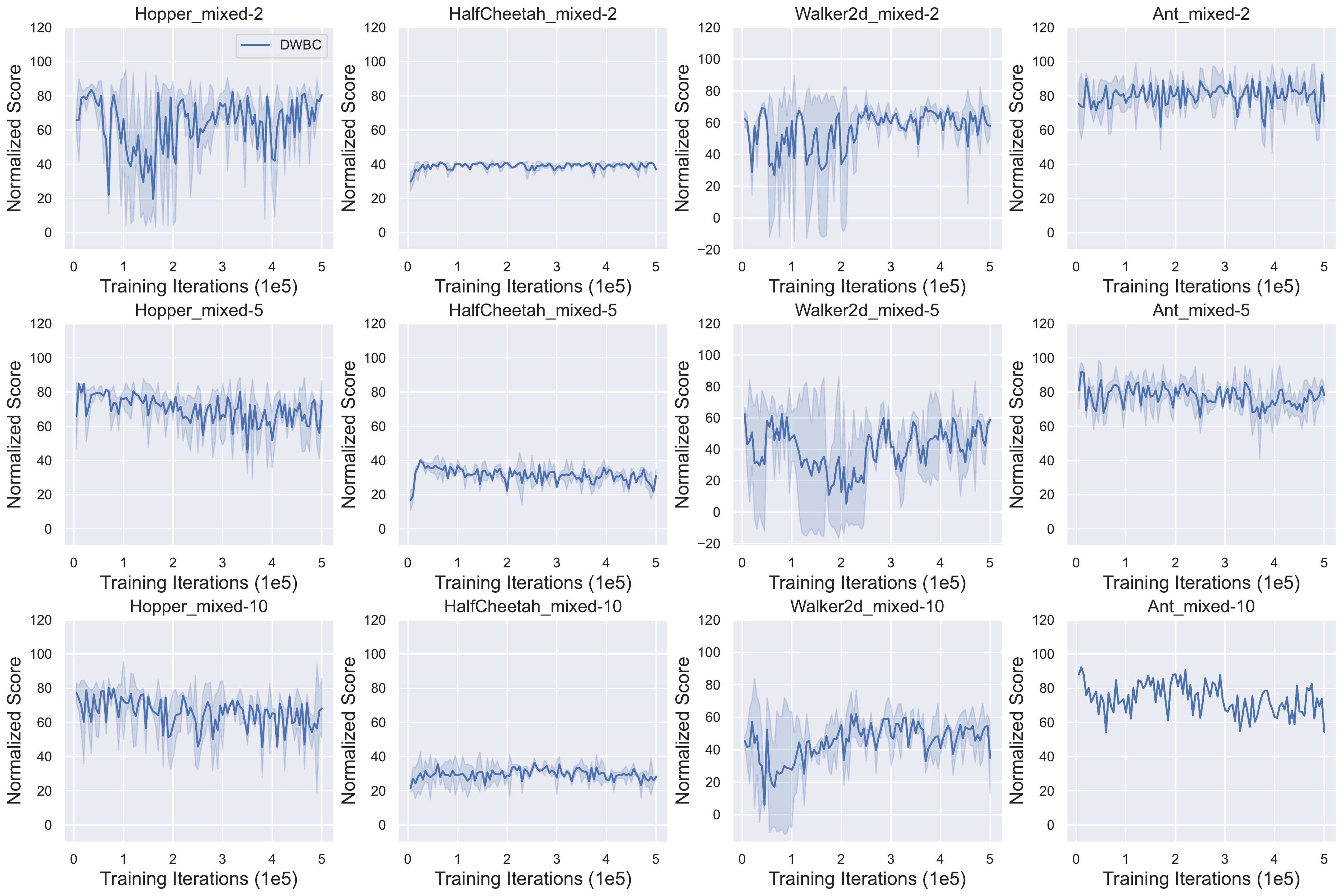} 
\caption{Learning curves for Setting 2.}
\label{fig_result2}
\end{figure*}

\begin{figure*}[b]
\centering
\includegraphics[width=0.9\columnwidth]{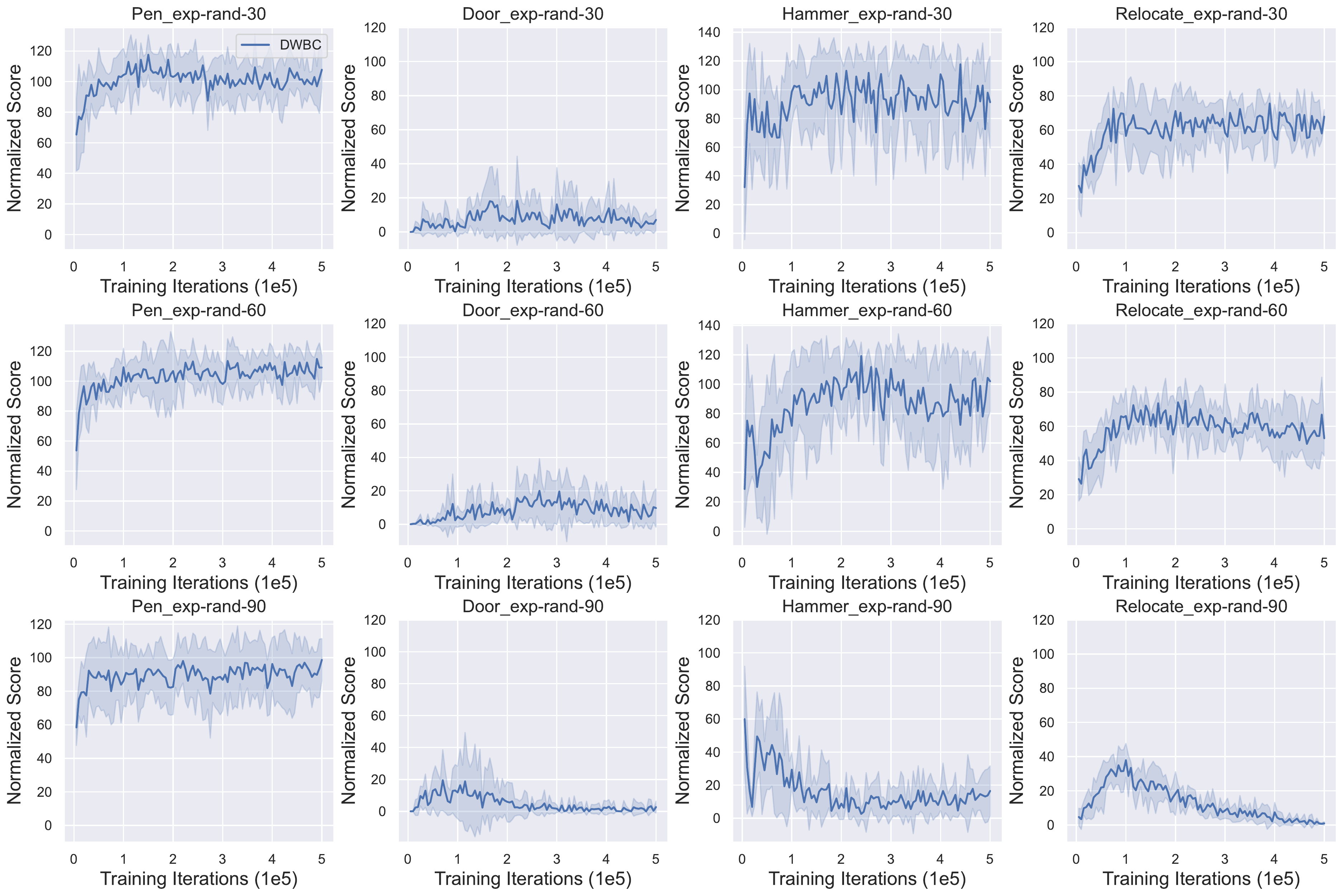} 
\caption{Learning curves for Setting 3.}
\label{fig_result3}
\end{figure*}

% In Figure \ref{fig_result}, we provide the learning curves of experiments conducted in Section \ref{sec:comp_eval}.
% As the learning procedure of BC is quite fast and stable, for more clearly presentation, we plot the results of BC-exp and BC-all as a horizon bar with the shaded area as the standard deviation across different seeds. 
% The average return in $\mathcal{D}_{e}$ is plotted as the red dashed line in all plots.
We provide the learning curves of DWBC in Figure \ref{fig_result1}, \ref{fig_result2} and \ref{fig_result3}.

% % \subsection{More comparison of DWBC and DWBC-old-d}
% % To more clearly see the comparison of DWBC and DWBC-old-d, we compute the mean value by dataset types.
% % It can be seen from Table \ref{table_comparison} that DWBC outperform DWBC-old-d by at least \textbf{10\%} on all type of datasets., we also find that DWBC achieves close to \textbf{20\%} improvement when $\mathcal{D}_o$ contains a large number of expert data.

% % \begin{table}[htb]
% % \centering
% % \caption{More comparison results.}
% % \vspace{5pt}
% % \begin{tabular}{c|c|c|c|c|c|c} 
% % \midrule
% % & \begin{tabular}[c]{@{}c@{}}\textbf{mixed-2}\\\textbf{mean}\end{tabular} & \begin{tabular}[c]{@{}c@{}}\textbf{mixed-5}\\\textbf{mean}\end{tabular} & \begin{tabular}[c]{@{}c@{}}\textbf{mixed-10}\\\textbf{mean}\end{tabular} & \begin{tabular}[c]{@{}c@{}}\textbf{\textbf{exp-rand-30}}\\\textbf{\textbf{mean}}\end{tabular} & \begin{tabular}[c]{@{}c@{}}\textbf{exp-rand-60}\\\textbf{mean}\end{tabular} & \begin{tabular}[c]{@{}c@{}}\textbf{exp-rand-90}\\\textbf{mean}\end{tabular}   \\ 
% % \midrule
% % DWBC-old-d                                                                      & 50.5    & 58.8    & 57.8    & 56.4    & 50.3     \\ 
% % \midrule
% % DWBC                                                                            & 57.3    & 67.0    & 67.9    & 62.3    & 61.6     \\ 
% % \midrule
% % \begin{tabular}[c]{@{}c@{}}Improvement \\(compared to DWBC-old-d)\end{tabular}  & 13.4\%  & 13.8\%  & 17.3\%  & 10.4\%  & 22.4\%.  \\
% % \midrule
% % \end{tabular}
% % \label{table_comparison}
% % \end{table}

\end{document}